\documentclass{article}
\usepackage{thumbpdf}
\usepackage{lmodern}
\usepackage{natbib}
\usepackage{hyperref}
\usepackage{longtable}
\usepackage{geometry}
\usepackage{pdflscape}
\usepackage{csvsimple}

\usepackage{xspace}
\usepackage{graphicx}
\graphicspath{{images/}}
\usepackage{amsmath, amssymb}
\usepackage{verbatim}
\usepackage[normalem]{ulem}
\usepackage{subfiles}
\usepackage[colorinlistoftodos]{todonotes}
\usepackage{multirow, makecell}

\newcommand{\Nbar}{$\bar{N}$\xspace}
\newcommand{\Nbarn}[1]{$\bar{N} = #1$}
\newcommand{\nuMatern}[1]{$\nu = \frac{#1}{2}$}
\newcommand{\thetat}{$\theta$\xspace}

\newcommand{\pusable}{$\delta_{0.05}$\xspace}

\newcommand*\vectheta{\boldsymbol{\theta}}
\renewcommand\vec{\mathbf}

\author{Timo Braun\\ Anders Kvellestad\\ Riccardo De Bin\\
University of Oslo, Norway}

\title{\texttt{GPTreeO}: An \texttt{R} package for continual regression with dividing local Gaussian processes}

\begin{document}

\maketitle

\begin{abstract}
We introduce \texttt{GPTreeO}, a flexible \texttt{R} package for scalable Gaussian process (GP) regression, particularly tailored to continual learning problems.
\texttt{GPTreeO} builds upon the \textit{Dividing Local Gaussian Processes} (DLGP) algorithm, in which a binary tree of local GP regressors is dynamically constructed using a continual stream of input data. 
In \texttt{GPTreeO} we extend the original DLGP algorithm by allowing continual optimisation of the GP hyperparameters, incorporating uncertainty calibration, and introducing new strategies for how the local partitions are created. Moreover, the modular code structure allows users to interface their favourite GP library to perform the local GP regression in \texttt{GPTreeO}. 
The flexibility of \texttt{GPTreeO} gives the user fine-grained control of the balance between computational speed, accuracy, stability and smoothness. 
We conduct a sensitivity analysis to show how \texttt{GPTreeO}'s configurable features impact the regression performance in a continual learning setting.
\end{abstract}

{\it emulator, local approximations, machine learning algorithm, online learning, online regression}

\vspace{2\baselineskip}\noindent

\section{Introduction}

Gaussian process (GP) regression is a well-known statistical/machine learning method widely used for data analysis and surrogate modeling. Thanks to its flexibility and its probabilistic nature, that allows a theoretically-sounded quantification of the uncertainty, it has been successfully applied in many fields, including chemistry \citep[see, e.g.,][]{FriederichAl2020}, physics \citep{Buckley:2020bxg}, and control applications \citep{ WangAl2021,PootAl2022}. 
One of the main drawbacks of GP regression is its computational cost, which constitutes a severe limitation in modern contexts characterised by large data collections. In particular, it requires the inversion of an $N \times N$ (covariance) matrix, an operation of order $O(N^3)$, where $N$ is the size of the data set. When the number of observations increases, the computations quickly become unfeasible. This becomes particularly problematic in the case of continual regression, when new data points are constantly added to the dataset.

To tackle this issue several approaches have been proposed in the literature; see \cite{LiuAl2020} for a complete review. They can be divided into three groups according to their strategy: (i) the global GP is approximated by using a lower-dimensional representation of the sample space \citep[as in, among many other,][]{CsatoOpper2002,McIntireAl2016,KoppelAl2021}; (ii) the matrix inversion step is sped up by mathematical or computational tricks \citep[see, e.g.,][]{Gneiting2002,AmbikasaranAl2015,AminfarAl2016}; (iii) local approximations are performed in a divide-and-conquer approach \citep[see, for example,][]{GramacyLee2008,DeisenrothNg2015,RulliereAl2018,10.1371/journal.pone.0256470}.

Among the methods that follow the latter strategy, \cite{LedererAl2020,pmlr-v139-lederer21a} proposed the Dividing Local Gaussian Processes (DLGP) approach.\footnote{In \cite{LedererAl2020} the approach is called Dividing Local Gaussian Processes (DLGP), while in \cite{pmlr-v139-lederer21a} it is referred to as Locally Growing Tree of GPs (LoG-GP). In this paper we adapt the terminology of \cite{LedererAl2020}.} The sample space is iteratively partitioned into regions in a manner of a statistical tree regression, and a single GP is associated with each partition. DLGP was developed with speed as a primary goal and consequently tested on online regression for moving robotic arms. 
\cite{LohrmannAl2022} adapted the algorithm to deal with online regression in global fits of new physics theories, giving up some speed to improve performance in terms of prediction error and to provide a better quantification of the uncertainty around the predicted values.
In addition to the important application in physics, the latter study showed how flexible the original DLGP can be made by removing some constraints due to specific choices made in the original algorithm. This paper presents our extended version of the DLGP algorithm, as implemented in a dedicated \texttt{R} package named \texttt{GPTreeO}.\footnote{See \url{https://CRAN.R-project.org/package=GPTreeO} and \url{github.com/timo-braun/GPTreeO}} We show how our multiple additions to the DLGP algorithm allow the user to control speed, accuracy, stability, smoothness, and uncertainty quantification. The algorithm can then be tuned to applications in multiple contexts and for multiple purposes. 

The rest of the paper is organised as follows: In Section \ref{method} we briefly review GP regression and its DLGP version. The \texttt{GPTreeO} package is described and demonstrated in Section \ref{implementation}. In Section \ref{sec: sensitivity analysis} we investigate what effect the different features in \texttt{GPTreeO} have on online learning performance. Finally, in Section \ref{sec: conclusions} we conclude and comment on possible directions for future work.

\section{Methods}\label{method}

\subsection{Background}

\subsubsection{Gaussian process regression}

A Gaussian process is a collection of variables for which the joint probability distribution for any finite subset of variables is a multivariate Gaussian distribution. We can use a GP as a Bayesian approach to regression if we let the variables in the GP represent the unknown values of some function $f(\vec{x})$ at different points $\vec{x}$. In this view a GP represents a probability distribution function (pdf) on the space of functions, and the starting point for regression is to define a prior GP that expresses our prior beliefs about the unknown $f(\vec{x})$. The mean and covariance matrix of this joint Gaussian pdf are specified by choosing a prior mean function $m(\vec{x})$ and a covariance function (or \textit{kernel}) $k(\vec{x},\vec{x}')$ such that $E[f(\vec{x})] = m(\vec{x})$ and $\mathrm{cov}(f(\vec{x}),f(\vec{x}')) = k(\vec{x},\vec{x}')$. 

The mean function is commonly just set to zero, while the choice of kernel is the crucial modelling step in GP regression. The kernel encodes prior expectations regarding the nature of the unknown function, such as smoothness, periodicity, etc. The training step in GP regression consists of optimising any free parameters in the kernel by maximising the marginal likelihood of the known function evaluations in the training data. Once we have a fully specified kernel and mean function, we condition our joint Gaussian pdf on the known function evaluations to obtain a Gaussian predictive posterior for the function value at other input points.

Let $X$ represent a matrix constructed from a set of $N$ input points, $X = [\vec{x}_1, \ldots, \vec{x}_N]^T$, and let $\vec{f}$ be a vector containing the corresponding function values, $\vec{f} = [f_1, \ldots, f_N]^T$, where we use the notation $f_i \equiv f(\vec{x}_i)$. Further, let $\vec{x}_*$ and $f_*$ denote a test input point and its corresponding function value. The predictive posterior for the function value at the test point then becomes the Gaussian distribution \citep{RasmussenCarlEdward2006Gpfm}
\begin{align}
  \label{eq:GP_predicted_f}
  p(f_* \big| X, \vec{f}, \vec x_*) = \phi(\mu_*, \sigma_*^2),
\end{align}
where $\phi(\mu_*, \sigma_*^2)$ is the probability density function of a normal distribution with mean and variance given by
\begin{align}
  \label{eq:GP_mean_predicted_f}
  &\mu_*=m(\vec x_*) + k(\vec x_*, X)\Sigma^{-1} (\vec f - m(X)), \\
  \label{eq:GP_variance_predicted_f}
  &\sigma_*^2=k(\vec x_*, \vec x_*)- k(\vec x_*, X)\Sigma^{-1}k(X, \vec x_*).
\end{align}
Here we have used the notation
\begin{align}
\Sigma        & \equiv k(X, X)\mathrm{,\ i.e.\ }\Sigma_{ij} = k(\vec x_i, \vec x_j) \label{eq:covar_matrix_def},\\
k(\vec x_*,X) &\equiv [k(\vec x_*,\vec x_i),\ldots,k(\vec x_*,\vec x_n)],\\
k(X,\vec x_*) &\equiv k(\vec x_*,X)^\mathrm{T}.
\end{align}

We also need to consider the case where our training data consist of a set of uncertain observations $y_i$ rather than the true function values $f_i$. If we model this as $y_i = f(\vec{x}_i) + \epsilon_i$, with $\epsilon_i \sim \mathcal{N} (0, \sigma_i)$, the predictive posterior for a new observation $y_*$ becomes 
\begin{align}
  \label{eq:GP_predicted_y}
  p(y_* \big| X, \vec{y}, \vec x_*) = \phi(\mu^\prime_*, \sigma_*^{\prime 2}),
\end{align}
where
\begin{align}
  \label{eq:GP_mean_predicted_y}
  &\mu^\prime_*=m(\vec x_*) + k(\vec x_*, X)[\Sigma + \text{diag}(\sigma^2_i)]^{-1} (\vec y - m(X)), \\
  \label{eq:GP_variance_predicted_y}
  &\sigma_*^{\prime 2} = k(\vec x_*, \vec x_*)- k(\vec x_*, X)[\Sigma + \text{diag}(\sigma^2_i)]^{-1} k(X, \vec x_*).
\end{align}
Here $\text{diag}(\sigma^2_i)$ is an $N \times N$ matrix with the individual $\sigma^2_i$ on its diagonal, and otherwise the notation is the same as in Eqs.\ (\ref{eq:GP_mean_predicted_f}) and (\ref{eq:GP_variance_predicted_f}).

In this work we will use three different, common choices for the kernel $k$: the squared exponential (or Gaussian) kernel and the Mat\'{e}rn kernel with two choices for the smoothness parameter $\nu$, namely $\nu = \frac{3}{2}$ and $\nu = \frac{5}{2}$. For a one-dimensional input the Gaussian kernel is given by
\begin{align}
  \label{eq:GP_Gaussien_kernel}
  k_\text{G}(x,x';l) = \exp\left(-\frac{(x-x')^2}{2l^2}\right),
\end{align}
where the length-scale hyperparameter $l$ determines how quickly $\mathrm{cov}(f(x),f(x'))$ decreases with increasing distance $|x-x'|$. The Mat\'{e}rn kernels with $\nu = \frac{3}{2}$ and $\nu = \frac{5}{2}$ are\footnote{
The general form of the Mat\'{e}rn kernel is
\begin{align}
  \label{eq:GP_Matern_kernel_general}
  k_\mathrm{M} (x, x'; \nu, l)
    = \frac{2^{1-\nu}}{\Gamma (\nu) } \Big( \sqrt{2 \nu}\frac{|x - x'|}{l} \Big)^{\nu} K_{\nu} \Big( \sqrt{2 \nu}\frac{|x - x'|}{l} \Big).
\end{align}
Here $\Gamma (\nu)$ is the gamma function and $K_{\nu}$ is the modified Bessel function of the second kind. 
}
\begin{align}
  \label{eq:GP_Matern_kernel_32}
  k_\text{M}^{\nu=3/2}(x,x';l) &= \left( 1 + \frac{\sqrt{3} |x-x'|}{l} \right) \text{exp} \left( - \frac{\sqrt{3} |x-x'|}{l} \right) ,\\
  \label{eq:GP_Matern_kernel_52}
  k_\text{M}^{\nu=5/2}(x,x';l) &=  \left( 1 + \frac{\sqrt{5} |x-x'|}{l} + \frac{5 (x-x')^2}{3 l^2} \right) \text{exp} \left( - \frac{\sqrt{5} |x-x'|}{l} \right).
\end{align}
Typically, the Mat\'{e}rn kernels are best used for rougher functions, while the Gaussian kernel is more suitable for smoother functions. When working with $d$-dimensional input points $\vec{x}$, we use an anisotropic, multiplicative kernel
\begin{align}
  k(\vec{x},\vec{x}';\sigma^2,\vec{l}) = \sigma^2 \prod_{j=1}^{d} k(x^{(j)}, x'^{(j)};l^{(j)}),
\end{align}
where $k(x^{(j)}, x'^{(j)};l^{(j)})$ is a one-dimensional kernel for component $j$ of $\vec{x}$, $\vec{l}$ is a vector of length scales $l^{(j)}$, and $\sigma^2$ is a scaling hyperparameter.

\subsubsection{Dividing Local Gaussian Processes}

The DLGP approach to continual GP regression proposed by \cite{LedererAl2020} is a divide-and-conquer strategy in which the stream of data points is used to dynamically partition the accumulated data set, and each partition is associated with a small GP model. The dynamic splitting of the ever-growing data set can be viewed as a growing tree structure, where the outermost nodes (the \textit{leaves}) each contain a subset of the data and a corresponding GP. Each new data point is assigned to a single leaf, as detailed below. When the number of points in a leaf reaches a set threshold $\bar{N}$, the leaf is split into two new leaf nodes, each with its own GP model, and the $\bar{N}$ data points are distributed to these two child nodes. To label the nodes we use a notation that explicitly identifies each node's position in the tree: $i = \texttt{0}$ denotes the root node; $i = \texttt{00}$ and $i = \texttt{01}$ are the first and second child nodes of the root node; $i = \texttt{000}$, $i = \texttt{001}$, $i = \texttt{010}$ and $i = \texttt{011}$ are the possible nodes two levels down, etc.

A probabilistic approach is used to assign each input point to a single leaf node: each branch point $i$ in the tree is associated with a probability function $p_{i\texttt{0}}(\vec{x})$, which sets the probability for an input point $\vec{x}$ to be assigned to the first of the two child nodes at this branch point. The probability $p_{i\texttt{1}}(\vec{x})$ for assignment to the second child node is then $p_{i\texttt{1}}(\vec{x}) = 1 - p_{i\texttt{0}}(\vec{x})$, and for the root node we simply have $p_{\texttt{0}}(\vec{x}) = 1$. By successive sampling from Bernoulli distributions with parameter $p = p_{i\texttt{0}}(\vec{x})$, a new data point can be propagated down the tree until it reaches a leaf node. 

\begin{figure}
    \centering
    \includegraphics[width=0.75\linewidth]{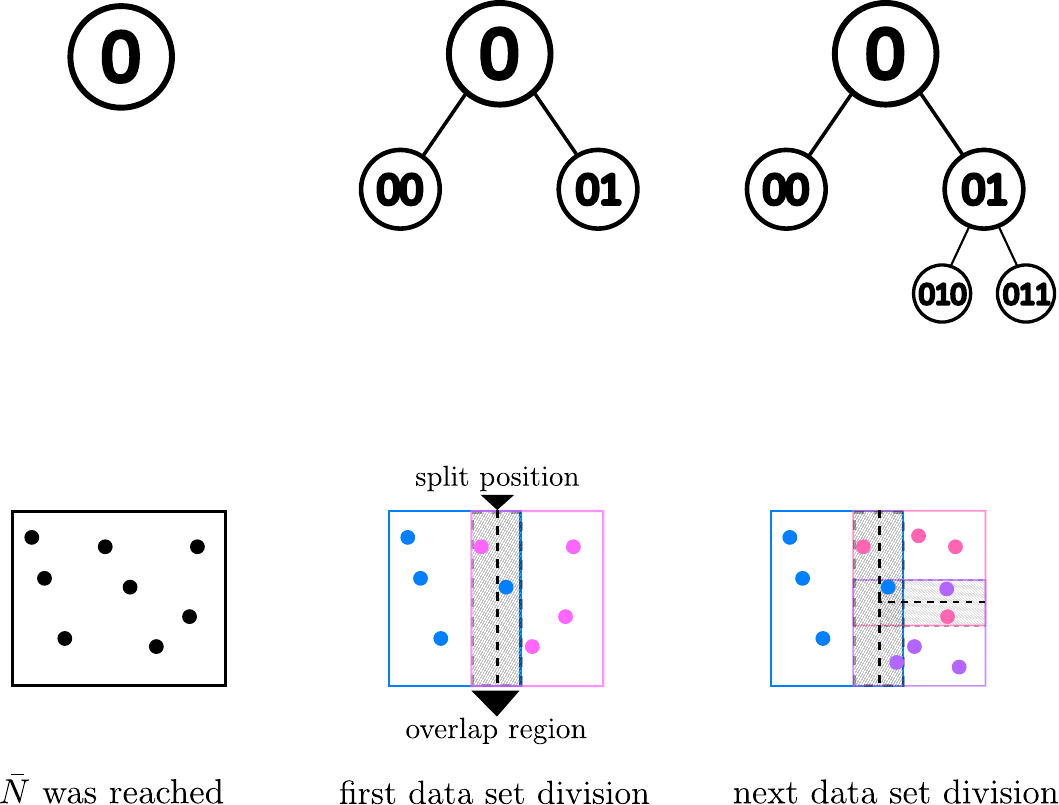}
    \caption{Visual representation of how the splitting works. The tree structure is shown above. Reproduction of Figure 1 from \cite{LedererAl2020}.}
    \label{fig: how splitting works}
\end{figure}

As discussed in \cite{LedererAl2020}, it is beneficial to choose the probability function $p_{i\texttt{0}}(\vec{x})$ such that the data set partitioning also induces a spatial partitioning of the input space. This can be used to ensure that only one or a small number of local leaf-node GPs must be evaluated to compute the model prediction at some new test point. \cite{LedererAl2020} use saturating linear functions of the form
\begin{align}\label{eq: p_i0}
p_{i\texttt{0}}(\vec{x}) = 
  \begin{cases}
    0  &\text{if } x_{j_i} < s_i - \frac{\theta w_{i,j}}{2}, \\
    \frac{x_{j_i} - s_i}{\theta w_{i,j}} + \frac{1}{2}  &\text{if } s_i - \frac{\theta w_{i,j}}{2} \leq x_{j_i} \leq s_i + \frac{\theta w_{i,j}}{2}, \\
    1  &\text{if } s_i + \frac{\theta w_{i,j}}{2} < x_{j_i}.
  \end{cases}
\end{align}
Here the following notation is used: $j_i$ denotes the coordinate that will be used to divide the input space of node $i$ in two; $s_i$ is the split position along this coordinate; $w_{i,j}$ is the largest distance along $x_{j_i}$ between two training points in node $i$; and $\theta$ is an \textit{overlap parameter}, $\theta \in [0,1]$. It controls the length $\theta w_{i,j}$ of the region along $x_{j_i}$ in which subsequent input points can be assigned to either of the two child nodes of node $i$. Figure \ref{fig: how splitting works} gives a visual representation of this data partitioning procedure.

The marginal probability $\tilde{p}_j(\vec{x})$ for an input point $\vec{x}$ to be assigned to leaf node $j$ is then given by the product of probabilities along the path to the leaf node. 
Using these marginal probabilities the predictive distribution for an unknown $y_*$ at input point $\vec{x}_*$ is
\begin{align}\label{eq: p_DLGP}
  p_{\textrm{DLGP}}(y_* | X, \vec{y}, \vec{x}_*) =\! \sum_{\textrm{leaves } j} \tilde{p}_j(\vec{x}_*)\,p_{\textrm{GP}_j}(y_* | X_j, \vec{y}_j, \vec{x}_*), 
\end{align}
where $p_{\textrm{GP}_j}(y_* | X_j, \vec{y}_j, \vec{x}_*)$ is the predictive distribution from the GP in leaf node $j$, which is informed by the data subset $(X_j, \vec{y}_j)$. The summation is carried out over all leaves. The mean and variance of $p_{\textrm{DLGP}}(y_* | X, \vec{y}, \vec{x}_*)$ are \citep[see][]{LedererAl2020}
\begin{equation}\label{eq: mu DLGP}
    \mu_\text{DLGP}(\vec{x}_*) = \sum_{\textrm{leaves } j} \tilde{p}_j(\vec{x}_*)\,\mu_j(\vec{x}_*) 
\end{equation}
and
\begin{equation}\label{eq: var DLGP}
    \sigma^2_\text{DLGP}(\vec{x}_*) = \sum_{\textrm{leaves } j} \tilde{p}_j(\vec{x}_*) \left( \sigma^2_j(\vec{x}_*) + \mu^2_j(\vec{x}_*) \right) - \mu^2_\text{DLGP}(\vec{x}_*), 
\end{equation}
where $\mu_j(\vec{x}_*)$ and $\sigma^2_j(\vec{x}_*)$ denote the mean and variance of $p_{\textrm{GP}_j}(y_* | X_j, \vec{y}_j, \vec{x}_*)$.

The overlap parameter $\theta$ plays the role of a smoothing parameter for the DLGP prediction. In the limit $\theta = 0$, all branching probabilities are either $p_\texttt{i\texttt{0}} = 0$ or $p_\texttt{i\texttt{0}} = 1$, and the predictive distribution in Eq.\ (\ref{eq: p_DLGP}) reduces to the distribution $p_{\textrm{GP}_{j'}}(y_* | X_j, \vec{y}_j, \vec{x}_*)$ of the single leaf GP $j'$ with marginal probability $\tilde{p}_{j'} = 1$. But for any non-zero choice of $\theta$, the DLGP prediction becomes a weighted sum of the predictions from all leaf nodes with $\tilde{p}_j(\vec{x}) > 0$ at the given input point $\vec{x}$.

\subsection{Novelties}

\subsubsection{Improved estimation of GP parameters}
A key difference between \texttt{GPTreeO} and the original DLGP algorithm of \cite{LedererAl2020} concerns how the parameters for each GP in the tree are estimated. In the original algorithm each new GP inherits the parameters from the GP in the root node. In contrast, the default in \texttt{GPTreeO} is that whenever a new leaf node is created, the parameters of its GP are estimated from the data contained in that node. While computationally slower, this choice ensures that each leaf contains a GP trained on the local behaviour of the target function.

To allow for different trade-offs between speed and accuracy, \texttt{GPTreeO} allows the user to set the frequency with which a GP should be retrained after the creation of a new leaf node. For this purpose we introduce a new parameter $b$, the \textit{retrain buffer}. It controls how many new points a node needs to receive before its GP is retrained with the new data. For instance, with settings \Nbar= 200 and $b$ = 15, new leaf GPs will start with on average 100 (= \Nbar/2) input points, and will be retrained for every fifteenth input point they receive.

An often useful balance between speed and accuracy is to only train each GP when it is first created and not retrain it later. Setting $b$ = \Nbar guarantees this behaviour, since even in the unlikely case where a leaf starts with only a single point, it will reach \Nbar points and split before reaching its retrain buffer limit. In Section \ref{sec: sensitivity analysis} we demonstrate this configuration for a tree with \Nbar = 1000. It is important to keep in mind that setting $b$ = \Nbar still means that new GPs on average will be trained with \Nbar/2, and not \Nbar, data points. The exception to this is when the new \textit{gradual splitting} feature discussed below is used, in which case each GP always contains exactly \Nbar points.

\subsubsection{Gradual splitting}\label{sec: gradual splitting}

\begin{figure}
    \centering
    \includegraphics[width=\linewidth]{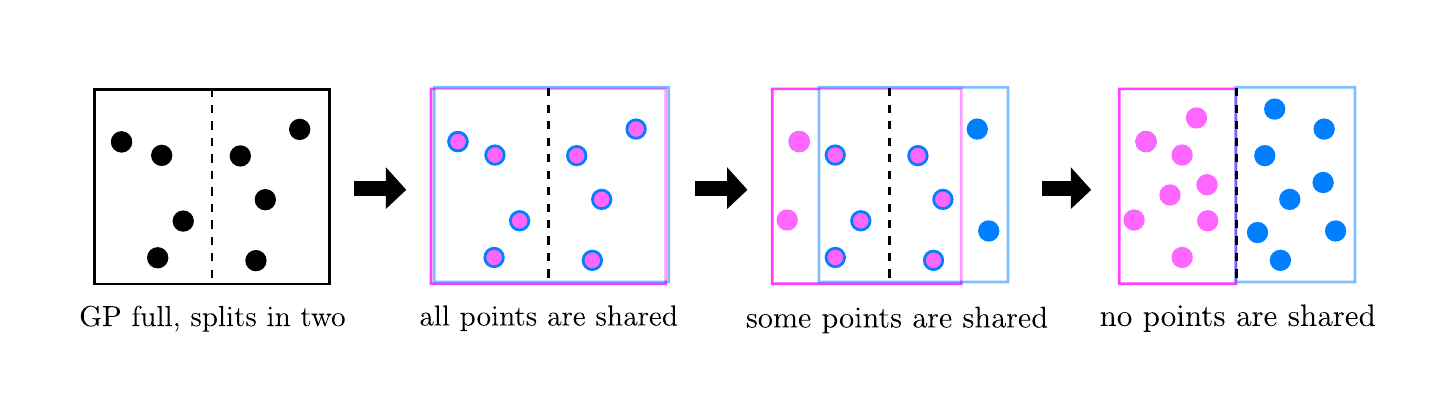}
    \caption{An illustration of the gradual splitting approach to creating new nodes. When a node splits, two identical child nodes are created that each inherit the full dataset from the parent node (\textit{first and second panel}). Each new data point is assigned to only one child node, and the chosen node simultaneously discards an old point that was shared with the other child node (\textit{third panel}). Eventually, the two child nodes have no data points in common (\textit{fourth panel}) and each can split into two new child nodes.}
    \label{fig: gradual splitting}
\end{figure}

In \texttt{GPTreeO} we introduce a new approach to node splitting, which we call \textit{gradual splitting}. It is visualised in Fig.\ \ref{fig: gradual splitting}: When a node reaches \Nbar training points, a split position is determined, two child nodes are created, and all \Nbar points are copied to each of the child nodes. Therefore, the child nodes start out as identical twins. Then, as usual, each new point is assigned to one of the child nodes as determined by the split position. When a new point is added to the left child node, its right-most point is deleted from its dataset and only remains in the right child node. Since the procedure works similarly when a new point is added to the right child node (the left-most point is removed), the two child nodes gradually drift apart as new points are added, until complete separation is achieved. During the process the total number of points in each node (\Nbar) remains constant.

Gradual splitting helps to ensure a more stable performance, since each leaf GP is always informed by \Nbar points. This differs from the original algorithm, where a child node starts with around $\bar{N}/2$ training points and gradually grows to \Nbar points. In this case, the predictions from a newly created child node are often less accurate than corresponding predictions from the parent node right before the split. On the other hand, since gradual splitting implies that each GP always has \Nbar points, the training and evaluation time for an individual GP will on average be longer compared to in the original algorithm with the same \Nbar value.

\subsubsection{Uncertainty calibration}\label{sec: uncertainty calibration}

A GP with a fully specified kernel represents a Bayesian degree of belief across the space of possible true functions. The GP prediction uncertainty for a given test point should therefore be interpreted in a Bayesian manner and not as a statement about coverage under repeated predictions at the same input point. However, in applications of emulators, it may be desirable to have prediction uncertainties that can be interpreted in terms of coverage. The repeated trial implicit in this notion of coverage would consist of receiving a new test point from the input stream and evaluating the emulator prediction for that point. The resulting distribution of residuals will therefore depend not only on the trained emulator, but also on the unknown and possibly evolving distribution that generates the stream of input points.

A common approach to uncertainty calibration is achieved through quantiles \citep[see][and references within]{SongAl2019}. A regression model is calibrated to the 68\% quantile if 68\% of the true values lie within the 68\% prediction interval. For \texttt{GPTreeO} we opt for a simple procedure, requiring only recent observations to fulfil the requirement mentioned above. This enables a continual calibration of the prediction uncertainty to an approximate 68\% coverage: when a new training point $(\vec{x}_i, y_i)$ is received and assigned to a GP leaf node, we first compute the corresponding prediction from this GP and store the residual ($e_i$) and the GP prediction uncertainty ($\sigma_i$). Each leaf node keeps track of its 25 most recent residuals and prediction uncertainties. From this we compute a single scaling factor $s$ by requiring $s \sigma_i \geq |e_i|$ for 68\% of the currently stored $(e_i,\sigma_i)$ pairs. When a node is full and splits into two child nodes, each child node inherits the last 25 $(e_i,\sigma_i)$ pairs and the corresponding $s$ value from the parent node to use as starting point for subsequent uncertainty calibration in the child node. The reason for calibrating against only the 25 most recent predictions, rather than e.g.\ the \Nbar latest predictions, is to ensure that the scaling factor can adapt quickly to changes in the distribution underlying the input stream. 
Since this calibration is performed locally for each node, the expected coverage is achieved when using $\theta = 0$, i.e., when each prediction is based on a single leaf GP. For $\theta > 0$ the effective coverage will typically be higher, though closer to 68\% than if the uncertainty calibration is not used.

\begin{figure}\centering
\includegraphics[width=\textwidth]{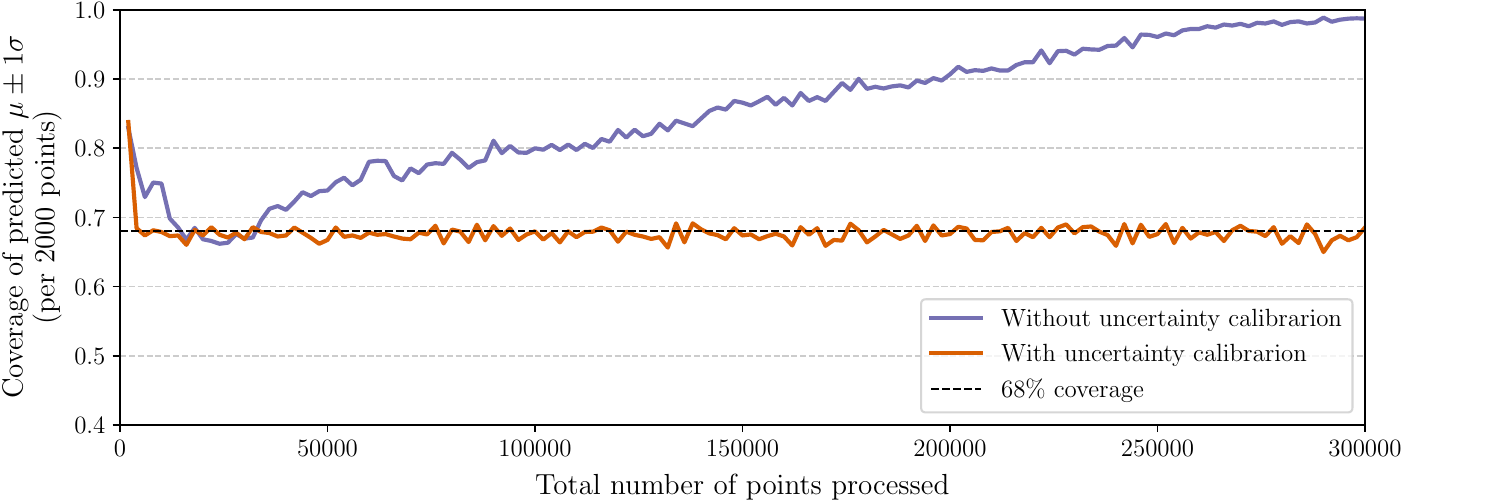}
\caption{\label{fig: uncertainty calibration} Coverage of the $\mu \pm 1\sigma$ prediction in a \texttt{GPTreeO} test run, computed for batches of 2000 input points. The graphs show the results with (orange) and without (purple) using \texttt{GPTreeO}'s uncertainty calibration option.}
\end{figure}
In Fig.\ \ref{fig: uncertainty calibration} we illustrate the impact of using this uncertainty calibration on a \texttt{GPTreeO} test run with \textit{test setup 1}, to be defined in Sec.\ \ref{sec: test setup}, and using $\theta = 0$. The plot shows the coverage computed for batches of 2000 input points, with and without uncertainty calibration. We see that, after a short period of initial training, the coverage for the calibrated prediction uncertainty remains approximately 68\%. To ensure conservative prediction uncertainties during the initial stages of the run, we initialise the scaling factor at a large value and limit the rate at which this estimate is allowed to decrease. 

\subsubsection{Additional features}\label{sec: hyperparameters}
In addition to the methodological advances described in the previous sections, \texttt{GPTreeO} includes further options so the approach can be better adapted to a wide range of problems. We will show how to choose these \texttt{GPTreeO} settings in Sec.\ \ref{sec: code settings}.

\paragraph{Covariance function.} The original algorithm only considers the squared exponential kernel, while our implementation, by allowing the user to connect their favourite GP implementation to \texttt{GPTreeO}, works with any covariance function supported by the given GP package. For example, when using \texttt{GPTreeO} with the \texttt{DiceKriging} package, one can use the Gaussian, the exponential, the power exponential, or the Matérn ($\nu = \tfrac{3}{2}$ or $\nu = \tfrac{5}{2}$) kernel.

\paragraph{Split direction.} In the original DLGP algorithm, the splitting of a GP node into two child nodes happens by dividing the node's data using the $\vec{x}$ coordinate for which the data spread is largest. In \texttt{GPTreeO} we provide the following options for how the split direction should be determined each time a node splits in two:

\begin{itemize}
    \item use the coordinate along which the node data has the largest spread -- this corresponds to the original DLGP algorithm;
    \item use the coordinate for which the length-scale hyperparameter of the node GP is smallest;
    \item use the coordinate for which the node data has the largest spread relative to the corresponding GP length-scale hyperparameter;
    \item use the coordinate which in the node data is most strongly correlated with the target $y$;
    \item use the first principal component.
\end{itemize}
The use of principal components-based splitting in a dividing GP approach has previously been studied in \cite{10.1371/journal.pone.0256470}. 

\paragraph{Split position.} The original DLGP algorithm sets the split point at the mean data position along the split direction, but \citet[Appendix B.1]{LedererAl2020} suggest that alternatives should be considered. In \texttt{GPTreeO} we also offer the option to use the median data position, as a more robust alternative to the mean.

\paragraph{Alternative probability functions for data partitioning.} In addition to the default function in Eq.\ (\ref{eq: p_i0}), where the probability $p_{i\texttt{0}}(\vec{x})$ decays linearly within the overlap region, we also allow for $p_{i\texttt{0}}(\vec{x})$ functions where the probability decays exponentially or as a half-Gaussian in the overlap region. In the code we refer to these choices as different ``decay shapes''.

\section{Implementation}\label{implementation}

\subsection{Outline of the \texttt{GPTreeO} package}

\texttt{GPTreeO} enables the user to create a tree-structured ensemble of local Gaussian processes, which functions in a similar way to a standard GP. The tree can be continually updated with new training data at low computational cost, making \texttt{GPTreeO} well suited for tackling streaming data and large data sets. The current version is aimed at applications where the input space is bounded, e.g.\ geo-spatial data or data from bounded optimisation problems, or where the input space only expands slowly relative to how quickly the GP tree evolves. Thus, we do not expect good performance for time series and similar applications where the input space is continually expanding in the main input coordinate (e.g.\ time).

Efficient memory usage is important for \texttt{GPTreeO}, since the expected input is a long-running stream of data. Thus, \texttt{GPTreeO} uses the \texttt{R6} class framework. In contrast to \texttt{S3} or \texttt{S4} classes, \texttt{R6} classes do not create copies of an object once an action or a method is performed on that object.

The nodes of the tree are stored in a hash, which helps speed up the evaluation of predictions thanks to the constant retrieval time. This is especially useful for long data streams which will produce a tree with a high number of nodes. To save memory, nodes that are no longer leaves have their local GP removed.

\subsection{Modularity}

\texttt{GPTreeO} is designed to be modular in the sense that the implementation of the extended DLGP algorithm is kept independent of what particular GP package is used for the GPs in the leaf nodes. This modularity is achieved by defining an interface layer that separates \texttt{GPTreeO}'s DLGP algorithm from the details of the individual GP packages. We refer to this interface layer as a ``wrapper'', as its task is to wrap existing GP packages with the interface expected by \texttt{GPTreeO}. The expected  interface is detailed in the source file \texttt{CreateWrappedGP.R}. Connecting \texttt{GPTreeO} to a new GP package is then a question of writing a single wrapper file that connects the interface to the relevant functionality of the given GP package. 

By default, \texttt{GPTreeO} uses the \texttt{DiceKriging} package from \cite{DiceKriging} for the leaf node GPs. The interface to \texttt{DiceKriging} is defined in the source file \texttt{WrappedDiceKriging.R}. As a proof of concept of \texttt{GPTreeO}'s modularity, we have also included an interface to the GP package \texttt{mlegp} \citep{DancikAl2008}, as defined in the file \texttt{Wrappedmlegp.R}, but we do not use \texttt{mlegp} for any of the tests in this paper. Moreover, a copy of the \texttt{mlegp} wrapper is provided in the file \texttt{WrappedGP.R}, intended as a starting point for users who want to connect other GP packages. A tutorial on how to do this is provided in the \texttt{GPTreeO} vignette.

\subsection{Basic example}

In this section we provide a minimal example of how to use \texttt{GPTreeO} for continual prediction and training given a stream of input data.
To keep the example below short and self-contained, it has three key simplifications compared to a realistic application: first, the true target function is known and inexpensive to evaluate. Second, the stream of input points comes from a simple for-loop over a pre-generated set of points, rather than from some external mechanism or algorithm. And third, the \texttt{GPTreeO} predictions are not utilised for anything, such as reducing how often the true target function is evaluated. 
We assume that the package has been successfully installed. The minimal example is then as follows:
\begin{verbatim}
R> library(GPTreeO)
R> set.seed(42)
R> X <- as.matrix(runif(100)) 
R> target <- function(x) list(y = sin(2*pi*x) + 0.2*sin(8*pi*x) + 0.1*rnorm(1), 
+    y_var = 0.1**2)
R> predictions <- NULL
R> gptree <- GPTree$new(Nbar=25, theta=0, retrain_buffer_length=1)
R> for (i in 1:nrow(X)) {
R>   x <- X[i,]
R>   predictions <- rbind(gptree$joint_prediction(x), predictions)
R>   target_output <- target(x)
R>   gptree$update(x, target_output$y, target_output$y_var) 
R> }
\end{verbatim}
We begin by uniformly sampling from $[0,1]$ a set of 100 $\textbf{x}$ points that will be used for the data stream, and we define an example 1D target function \verb|target| that returns both a central value \verb|y| and a variance \verb|y_var| for a given input point. For the example target we use a similar 1D function as used in \cite{Higdon2002}, including a small noise term. Then, we create a new tree (an instance of the \verb|GPTree| class) with the call to \verb|GPTree$new()|. The example input arguments to \verb|new| specify that this tree should use $\bar{N}=25$ and $\theta=0$, and that leaf-node GPs should be retrained every time they are assigned a new training point.

Once the tree is created, we start the sequence of iterative prediction and training for the example input stream (the for-loop over the points stored in \verb|X|). For each new input point \verb|x|, we first predict the target value using the method \verb|joint_prediction|. This method returns the tree's prediction and prediction uncertainty, as given by Eqs. (\ref{eq: mu DLGP}) and (\ref{eq: var DLGP}). We then evaluate the true target function. Finally, we use the \verb|update| method to update the tree with the now-known target value for this input point. We note that the \verb|update| method expects a variance, not an uncertainty, as its third input argument. 

\subsection{Code settings}\label{sec: code settings}
In the following, we discuss the important settings when initialising a new tree using \verb|GPTree$new()|. A complete overview can be found in the documentation of the \texttt{GPTree} class.

\begin{itemize}
    \item \texttt{Nbar}. Maximum number of data points for each GP in a leaf before it is split. The default value is 1000.
    \item \verb|retrain_buffer_length|. Size of the retrain buffer $b$. The buffer for a each node collects data points and holds them until the buffer length is reached. Then, the GP in the node is updated with the data in the buffer. For a fixed \verb|Nbar|, higher values for \verb|retrain_buffer_length| lead to faster run time (less frequent retraining), but the trade-off is a temporary reduced prediction accuracy. We advise that the choice for \verb|retrain_buffer_length| should depend on the chosen \texttt{Nbar}. By default \verb|retrain_buffer_length| is set equal to \verb|Nbar|.
    \item \verb|gradual_split|. If \texttt{TRUE}, gradual splitting as discussed in Sec.\ \ref{sec: gradual splitting} is used for splitting. The default value is \texttt{TRUE}.
    \item \texttt{theta}. It corresponds to \thetat, the overlap ratio between two leaves in the split direction. The default value is 0.
    \item \verb|wrapper|. A string that indicates which GP implementation should be used. The current version includes wrappers for the packages \texttt{"DiceKriging"} and \texttt{"mlegp"}. The default setting is \texttt{"DiceKriging"}.
    \item \verb|gp_control|. A \texttt{list()} of control parameters that are forwarded to the wrapper. Here the covariance function is specified. \texttt{DiceKriging} allows for the following kernels, passed as string: \texttt{"gauss"}, \verb|"matern5_2"|, \verb|"matern3_2"|, \texttt{"exp"}, \texttt{"powexp"}, where \verb|"matern3_2"| is set as default.
    \item \verb|split_direction_criterion|. A string that indicates which splitting criterion to use. The options are:
    \begin{itemize}
        \item \verb|"max_spread"|: split along the direction which has the largest data spread.
        \item \verb|"min_lengthscale"|: split along the direction with the smallest length-scale hyperparameter from the local GP.
        \item \verb|"max_spread_per_lengthscale"|: split along the direction with the largest data spread relative to the corresponding GP length-scale hyperparameter.
        \item \verb|"max_corr"|: split along the direction where the input data is most strongly correlated with the target variable.
        \item \verb|"principal_component"|: split along the first principal component.
    \end{itemize}
    The default value is \verb|"max_spread_per_lengthscale"|.
    \item \verb|split_position_criterion|. A string indicating how the split position along the split direction should be set. Possible values are \texttt{"median"} and \texttt{"mean"}. The default is \texttt{"median"}.
    \item \verb|shape_decay|. A string specifying how the probability function $p_{i\texttt{0}}(\vec{x})$ should fall off in the overlap region. The available options options are a linear shape (\texttt{"linear"}), an exponential shape (\texttt{"exponential"}), or a Gaussian shape (\texttt{"gaussian"}). Another option is to select no overlap region. This can be achieved by selecting \texttt{"deterministic"} or to set \texttt{"theta"} to 0. The default is \texttt{"linear"}.
    \item \verb|use_empirical_error|. If \texttt{TRUE}, the uncertainty is calibrated using recent data points, as described in Sec.\ \ref{sec: uncertainty calibration}. The default value is \texttt{TRUE}.
\end{itemize}

\section{Sensitivity analysis}\label{sec: sensitivity analysis}

In this section we investigate how different choices for the tree and GP hyperparameters influence the performance of \texttt{GPTreeO}.  We will consider the hyperparameters' impact on the following indicators:
\begin{itemize}
    \item the average prediction error (RMSE);
    \item the fraction of the last 2000 input points for which the relative prediction error is $<5\%$ (\pusable);
    \item the average GP tree prediction uncertainty;
    \item the tree update time;
    \item the prediction time.
\end{itemize}
We include \pusable as a performance metric, despite its similarity with RMSE, because it more directly expresses the potential usefulness of \texttt{GPTreeO} as a surrogate model in applications with a certain error tolerance. 

To keep the terminology simple, we will use \textit{tree hyperparameters} (or the shorthand \textit{tree parameters}) to collectively refer to the kernel function choice for GPs  and the settings/parameters that control the tree construction. In our test runs we investigate all combinations of the following tree hyperparameter choices:
\begin{itemize}
    \item (\Nbar, $b$): (100, 2), (200, 15), (1000, 1000);
    \item \textit{overlap}: \thetat = 0.01, \thetat = 0.05, \thetat = 0.10 without gradual splitting; \thetat = 0 with gradual splitting;
    \item \textit{kernel type}: Gaussian, Matérn (\nuMatern{3}), Matérn (\nuMatern{5});
    \item \textit{split direction criterion}: maximum spread, maximum spread per length scale, first principal component.
\end{itemize}
We choose the above combinations of \Nbar and $b$ as they give comparable overall run times in our tests, but still serve to illustrate the difference between frequently retraining each GP and training each GP only when it is first created.
To avoid a too high number of different test configurations, we only test gradual splitting in combination with zero overlap ($\theta = 0$).
In all test runs we use the median to set the split position. Our preliminary tests indicated that the choice between mean and median had a very limited influence on all the indicators we consider in this analysis. But using the median has the small advantage that, for $\theta \rightarrow 0$, a parent node is guaranteed to distribute its data points evenly across the two child nodes. 
Lastly, since trees with the $(\bar{N}, b) = (1000, 1000)$ configuration do not train their first GP until the $1000^\text{th}$ input point is reached, we leave out the first 1000 points when computing the indicators above for all tested combinations of tree hyperparameters. 

For our analysis we rank the trees according to the indicator that we currently consider and investigate the frequency with which different hyperparamter settings occur at the top or bottom of the table. Once the most impactful features have been discovered, we consider remaining hyperparameters by conditioning the table on the previous features, separating the table into smaller ones. 
Our main set of test runs involves a total of 108 different \texttt{GPTreeO} configurations, due to the combinatorics of the different tree hyperparameter settings. 
For clarity, we will therefore limit our discussion to summarising our results and highlighting which tree parameters most strongly influence how \texttt{GPTreeO} performs. The complete set of test results is provided in the Appendix.

\subsection{Test setup}\label{sec: test setup}

A typical use case for \texttt{GPTreeO} is to learn a fast surrogate for an expensive computation $f(\vec{x})$, which is required as part of the numerical optimisation or exploration of some loss function $L(\vec{x})$. Therefore, a test run of \texttt{GPTreeO} needs two external ingredients, namely a true function $f(\vec{x})$ and a stream of input points $\vec{x}$. For each new point $\vec{x}$ in the input stream, we first let \texttt{GPTreeO} compute its corresponding prediction $\hat{f} \pm \Delta \hat{f}$. Then, the true target value $f(\vec{x})$ is evaluated and used to update the tree in preparation for the next input point in the stream.

Inspired by such use cases, we construct two main test setups:
\paragraph{Test setup 1:}
\begin{itemize}
    \item \textit{Input stream}: the sequence of $\vec{x}$ points sampled by a differential evolution optimiser exploring a four-dimensional Rosenbrock function \citep{10.1093/comjnl/3.3.175}, $L_\textrm{4D}(\vec{x})$. 
    \item \textit{Target function}: a four-dimensional Eggholder function \citep{WHITLEY1996245}, $f_\textrm{EH,4D}(\vec{x})$.
\end{itemize}
\paragraph{Test setup 2:}
\begin{itemize}
    \item \textit{Input stream}: the sequence of $\vec{x}$ points sampled by a differential evolution optimiser exploring an eight-dimensional Rosenbrock function, $L_\textrm{8D}(\vec{x})$. 
    \item \textit{Target function}: an eight-dimensional Robot Arm function \citep{AN2001588}, $f_\textrm{RA,8D}(\vec{x})$.\\
\end{itemize}
As an $n$-dimensional generalisation of the Rosenbrock function, we use
\begin{equation}
\label{eq: nD Rosenbrock function}
    L_\textrm{$n$D}(\vec{x}) = \sum\limits_{i=1}^{n-1} L_\textrm{2D}(x_i, x_{i+1}), 
\end{equation}
where each term $L_{2D}(x_i, x_{i+1})$ is a regular, two-dimensional Rosenbrock function,
\begin{equation}
    L_\textrm{2D}(x_i, x_{i+1}) = (1-x_i)^2 + 100 (x_{i+1} - x_i^2)^2,\quad x_i,x_{i+1} \in [-5,10].
\end{equation}
Similarly, we construct our four-dimensional version of the Eggholder function as
\begin{equation}
    f_\textrm{EH,4D}(\vec{x}) = \sum\limits_{i=1}^{3} f_\textrm{EH,2D}(x_i, x_{i+1}), 
\end{equation}
where
\begin{align}
\begin{split}
    f_\textrm{EH,2D}(x_i, x_{i+1}) = &-(x_{i+1} + 47)\sin\!\left(\sqrt{\left| \tfrac{1}{2}x_i + (x_{i+1} + 47) \right|}\right)\\ 
    &- x_i \sin\!\left(\sqrt{|x_i - (x_{i+1} + 47)|}\right) + 960.6407
\end{split}
\end{align}
for $x_i,x_{i+1} \in [-512,512]$. The last term in $f_\textrm{EH,2D}$ is simply a shift that ensures $f_\textrm{EH,2D}(x_i, x_{i+1}) \geq 1$. Finally, the eight-dimensional Robot Arm function is given by
\begin{equation}
\label{eq: RA function}
    f_\textrm{RA,8D}(\vec{x}) = 1000\,g(\vectheta, \vec{L}) + 1,
\end{equation}
with 
\begin{equation}
    g(\vectheta, \vec{L}) = \sqrt{\left[ \sum\limits_{i=1}^{4} L_i \cos\!\left( \sum\limits_{j=1}^{i} \theta_i \right) \right]^2 + \left[ \sum\limits_{i=1}^{4} L_i \sin\!\left( \sum\limits_{j=1}^{i} \theta_i \right) \right]^2},
\end{equation}
for $\theta_{1,2,3,4} \in [0,2\pi]$ and $L_{1,2,3,4} \in [0,1]$. The scaling and shift in Eq.\ (\ref{eq: RA function}) ensures that we work with $f_\textrm{RA,8D}$ values of $\mathcal{O}(1)$ rather than $\mathcal{O}(10^{-3})$ and that $f_\textrm{RA,8D} \geq 1$. Since for our test setups we want the target function and loss function to have the same domain, we scale all input parameters to $x_i \in [0,1]$.

For all \texttt{GPTreeO} runs with test setup 1, we use the same input stream of 4D $\vec{x}$ points, recorded from a single run of a differential evolution optimiser on $L_\textrm{4D}(\vec{x})$. Similarly, all runs with test setup 2 also use the exact same input stream of 8D $\vec{x}$ points. Each recorded input stream has a total of $3 \times 10^5$ points. However, due to the large number of different hyperparameter combinations we investigate, we stop the run in most of our \texttt{GPTreeO} test runs after the first $5 \times 10^4$ input points.

In both test setups above, we use an adaptive sampling algorithm (differential evolution) to generate a stream of input points $\vec{x}$.\footnote{The input streams were generated with the \texttt{DEoptim} package \citep{JSSv040i06,RJ-2011-005}, running with a population of 1000 members for 300 iterations.} This emulates the realistic use case where \texttt{GPTreeO} is used to learn a surrogate on the fly during some numerical optimisation or sampling task. Adaptive algorithms generally focus the computational effort on the most relevant sub-regions of the input space. For a surrogate to be useful in such applications, it must perform very well in the local regions of $\vec{x}$ space that the adaptive algorithm focuses on, while the surrogate's performance in other input regions is much less important. This is the reason why using a continually trained local surrogate like \texttt{GPTreeO} can be highly cost effective in such optimisation/sampling applications, compared to the alternative of training a global surrogate before starting the actual optimisation/sampling task. However, the use of adaptive sampling to generate our test data streams means that care must be taken when we interpret the time evolution of \texttt{GPTreeO}'s performance in our tests. While \texttt{GPTreeO} continually learns the target function from the input stream, it is simultaneously the case that the target function itself effectively becomes less challenging to learn (i.e.\ more slowly varying) as the adaptive sampling algorithm focuses in on a sub-region of the total input space. To explore this we perform some additional test runs where the input stream is generated by sampling $\textbf{x}$ points from a uniform distribution on the full input space. Lastly, we also carry out test runs where we add Gaussian noise to the target functions.

\subsection{Influence of tree parameters for test setup 1 (4D)}\label{sec: 4D test results}

The complete set of results from the runs are given in Tables \ref{tab: results trees 8000000}, \ref{tab: results trees 11000000} and \ref{tab: results trees 12000000} in the Appendix. We have removed six outliers, i.e.\ data points whose predictions are outside of [-100000, 100000]. For reference, all true values lie within [1000, 5000] for test setup 1.
By default, \texttt{DiceKriging} uses a constant GP mean function determined by a numerical optimization, and the six outliers correspond to cases where this optimization has picked out a suboptimal value. This issue can be solved by setting the mean function before calling \texttt{DiceKriging}, but this would go against our choice of delegating all GP calculations to the chosen external GP package.

\begin{table}[]
\begin{center}
\begin{small}
\begin{tabular}{lccccc}
\hline
Impact & RMSE & \pusable & Pred.\ uncert.\ & $t_{\text{update}}$ & $t_{\text{pred}}$ \\ 
\hline
\multirowcell{2}[0ex][l]{High} & \multirowcell{2}{(\Nbar, $b$) \\ kernel} & \multirowcell{2}{(\Nbar, $b$) \\ kernel} & \multirowcell{2}{\thetat} & \multirowcell{2}{(\Nbar, $b$)\\ \thetat} & \multirowcell{2}{(\Nbar, $b$)\\ \thetat} \\
 &  &  &  &  &  \\
 &  &  &  &  &  \\
\multirowcell{2}[0ex][l]{Medium} & \multirowcell{2}{\thetat} & \multirowcell{2}{\thetat} & \multirowcell{2}{split dir. \\ (\Nbar, $b$)} & \multirowcell{2}{kernel} &  \\
&  &  &  &  &  \\
&  &  &  &  &  \\
\multirowcell{2}[0ex][l]{Low} & \multirowcell{2}{split dir.\ } & \multirowcell{2}{split dir.\ } & \multirowcell{2}{kernel} &  \multirowcell{2}{split dir.\ }  & \multirowcell{2}{split dir.\ \\ kernel} \\
&  &  &  &  &  \\
\hline
\end{tabular}
\end{small}
\caption{Overview of the importance of tree parameters on performance indicators for test setup 1, the 4D data set. The more important a parameter, the higher up in the table it is, also within each row. The column in the middle refers to the un-scaled prediction uncertainty.}
\label{tab: importance parameters target quantities 4D}
\end{center}
\end{table}

\subsubsection{RMSE}\label{sec: RMSE 4D}
For test setup 1 we find that the RMSE is most affected by the choice of \Nbar and retrain buffer length. Trees using the combination $(\bar{N}, b) = (200, 15)$ have on average the lowest RMSE. This holds true irrespective of the kernel used. For trees using a Matérn kernel, larger \Nbar are preferred over smaller ones. In the case of the Gaussian kernel, the opposite holds: trees using $(\bar{N}, b) = (1000, 1000)$ have a higher average RMSE. 

Of similar magnitude is the effect of the choice of kernel type. Choosing a Matérn kernel improves the RMSE compared to using a Gaussian kernel. This is expected; the target function in this test (the 4D Eggholder function) is very quickly varying on the input space, and in such cases GPs with Matérn kernels will often capture local features of the target function more easily than GPs with Gaussian kernels.\footnote{The Gaussian kernel encodes a strong prior belief in very smooth functions. Concretely, sample functions from a GP with a Gaussian kernel will be infinitely differentiable. In comparison, a Matérn kernel with $v > k$ only ensures that the sample functions are $k$ times differentiable; see \cite{RasmussenCarlEdward2006Gpfm}.} 
Further comparing the Matérn kernels, trees with \nuMatern{3} have a lower RMSE than those with \nuMatern{5}. To add to our previous observations, we find that the difference in RMSE for different kernel choices becomes more pronounced with increasing \Nbar.

The overlap parameter \thetat also has some influence on the RMSE in the test runs. For \Nbarn{100}, choosing $\theta = 0.01$ is associated with an increased RMSE, while this effect is smaller for \Nbarn{200} and \Nbarn{1000}. The RMSE values for trees using gradual splitting are similar to those for trees with $\theta = 0.05$ or $0.1$.
The choice of method for determining the split direction has no noteworthy influence on the RMSE. 

\subsubsection{Fraction of test points with prediction error below 5\% (\pusable)}
As expected, the impact of the different tree hyperparameters on \pusable is closely linked to their impact on RMSE: in general, the parameter choices that lead to lower RMSE also lead to higher \pusable. For this reason, we only highlight the differences with respect to the discussion in the previous section.

One difference is that there is no longer a clear performance difference between combinations $(\bar{N}, b) = (100, 2)$ and $(1000, 1000)$ when using Matérn kernels. In every case, Matérn kernels outperform the Gaussian kernel in terms of \pusable. Furthermore, while $\theta = 0.01$ was associated with an increased RMSE, this choice does not seem to have a particular negative impact the \pusable performance. We also find that among the tree configurations with worst \pusable performance, the setting $\theta = 0.05$ is almost completely absent. This effect is relevant for trees using a Matérn kernel with \nuMatern{5} or a Gaussian kernel. 

\subsubsection{Prediction uncertainty}\label{sec: uncert estimate 4D}

We now turn to how the choice of tree hyperparameters influence the size of the prediction uncertainty, $\sigma_{\textrm{DLGP}}$, expressed by the variance in Eq.\ (\ref{eq: var DLGP}). However, for most practical applications of \texttt{GPTreeO}, we would recommend using the automatic uncertainty calibration discussed in Sec.\ \ref{sec: uncertainty calibration}. In this case the uncertainties are calibrated with a separate scaling factor for each leaf, meaning that the other tree parameters will mainly just influence the relative change in prediction uncertainty as function of $\vec{x}$ within each leaf node. 

The tree parameter \thetat has the biggest influence on the average prediction uncertainty in our test runs with test setup 1. Lowering \thetat, that is decreasing the overlap between leafs, yields an on average lower prediction uncertainty. The tree configurations that produce the most confident predictions in these test runs are those that use gradual splitting (with $\theta = 0$) and $(\bar{N}, b) = (100, 2)$ or $(200, 15)$. We remind the reader that the gradual splitting approach ensures that each leaf GP always operates with its maximum number of data points (\Nbar), which partly explains why this configuration gives the lowest prediction uncertainties. 

We find that trees splitting on the first principal component generally have higher uncertainty estimates. This effect is strongest for trees with $\theta = 0.1$. The choice of kernel function has a limited impact on the prediction uncertainty.  

\subsubsection{Update time}\label{sec: 4d training time}
As per design of the algorithm, we see in our test runs that the average time required to update the tree with a new training point is mainly determined by the choice of $(\bar{N}, b)$: higher \Nbar leads to a longer update time, higher $b$ lowers the update time.
The combinations $(100,2)$, $(200,15)$, and $(1000,1000)$ which we use in our tests were chosen because they are expected to give run times of the same order of magnitude, at least when gradual splitting is not used (i.e.\ the configurations $\theta = 0.01, 0.05, 0.10$ in our tests). Of the tested $(\bar{N}, b)$ configurations, we observe that $(100,2)$ results in the longest average update times. Further, using gradual splitting also significantly increases the average update time, as expected per the discussion in Sec.\ \ref{sec: gradual splitting}. For trees that use both $(\bar{N}, b) = (100,2)$ and gradual splitting, we find update times up to an order of magnitude higher than for the fastest trees in this test.

The other parameter choices have much smaller impact on the update time. In these test runs we note that for trees with $(\bar{N}, b) = (1000,1000)$, using one of the Matérn kernels gives shorter update times compared to using the Gaussian kernel. Presumably, the GP hyperparameter tuning typically requires fewer steps when using this Matérn kernel, which more easily adapts to the quickly varying test function.\footnote{The \texttt{DiceKriging} package is interfaced to the \texttt{optim} and \texttt{gencloud} packages for GP hyperparameter tuning. In our tests we use the BFGS optimiser from the \texttt{optim} function from the \texttt{stats} package \citep{RCoreTeam}.} When we switch off gradual splitting and instead use the original splitting approach with \thetat values 0.01, 0.05, and 0.10, we observe almost no difference between them, as expected. The same also holds for different splitting direction criteria.

\subsubsection{Prediction time}\label{sec: 4d prediction time}
Since evaluating a GP prediction has complexity $\mathcal{O}(N^2)$ for a GP with $N$ data points, the expected complexity for evaluating a prediction in the DLGP approach is $\mathcal{O}(n_\textrm{leaves} \bar{N}^2)$, where $n_\textrm{leaves}$ is the number of leaves contributing to the prediction for the given input point. Thus, the tree hyperparameters that should most strongly influence the average prediction time are \Nbar, which bounds the per-leaf complexity, and \thetat, which indirectly determines how fast the number of contributing leaves increases as the tree grows. 

This is what we observe in our test runs. The trees with $\bar{N} = 1000$ have the longest prediction times, and prediction times grow with increasing \thetat. For the trees where we combine gradual splitting and $\theta = 0$, the $\theta = 0$ choice ensures that only a single leaf is evaluated per prediction, but at the same time the gradual splitting means that the given leaf will always contain its maximum number of data points. In our test runs, where each tree receives a stream of $5\times10^4$ input points, we observe that using gradual splitting with $\theta = 0$ has similar impact on the average prediction time as using $\theta = 0.01$ with standard splitting, when using $(\bar{N}, b) = (100,2)$ or ($200, 15)$. For $(\bar{N}, b) = (1000,1000)$, gradual splitting yields prediction times similar to $\theta = 0.05$. However, for use-cases with very long-running data streams, one should keep in mind that all tree configurations that have a non-zero \thetat will have a number of contributing leaves, $n_\textrm{leaves}$, that slowly increases as the tree grows. As discussed in \cite{LedererAl2020}, when using probability functions that spatially partition the data, along with sufficiently small \thetat, the DLGP prediction time will for many applications grow only logarithmically with the total number of training points the tree has received. 
As expected, we observe that the prediction time is not influenced by the choice of kernel type or split direction criterion.

\subsection{Influence of tree parameters for test setup 2 (8D)}\label{sec: 8D test results}

For test setup 2 the stream of input $\vec{x}$ points are from a differential evolution exploration of the 8D Rosenbrock function in Eq.\ \eqref{eq: nD Rosenbrock function}, while the target function is the 8D Robot Arm function in Eq.\ \eqref{eq: RA function}. Given our choice of anisotropic Gaussian and Matérn kernels for our tests, training a single GP on an 8D test function implies optimising eight length-scale hyperparameters, compared to only four when testing with the 4D function of test setup 1. In practice this leads to average GP training times that are an order of magnitude higher for test setup 2. Due to this, in combination with the large number of test runs, we only consider the $(\bar{N}, b)$ configurations $(200,15)$ and $(1000,1000)$ for test setup 2, leaving out the $(100,2)$ configuration. This leaves us with 72 tree parameter combinations. The complete set of results from the runs are given in Tables \ref{tab: results trees 15000000} and \ref{tab: results trees 7000000} in the Appendix.

\begin{table}[]
\begin{center}
\begin{small}
\begin{tabular}{lccccc}
\hline
Impact & RMSE & \pusable & Pred.\ uncert.\ & $t_{\text{update}}$ & $t_{\text{pred}}$ \\ 
\hline
\multirowcell{2}[0ex][l]{High} & \multirowcell{2}{(\Nbar, $b$)} & \multirowcell{2}{kernel} & \multirowcell{2}{split dir.\  } & \multirowcell{2}{(\Nbar, $b$) \\ \thetat } & \multirowcell{2}{(\Nbar, $b$) \\ \thetat} \\
 &  &  &  &  &  \\
 &  &  &  &  &  \\
\multirowcell{2}[0ex][l]{Medium} & \multirowcell{2}{split dir.\ \\ kernel} & \multirowcell{2}{(\Nbar, $b$) \\ \thetat } & \multirowcell{2}{\thetat} & \multirowcell{2}{kernel} &  \\
&  &  &  &  &  \\
&  &  &  &  &  \\
\multirowcell{2}[0ex][l]{Low} & \multirowcell{2}{\thetat } & \multirowcell{2}{split dir.\ } & \multirowcell{2}{(\Nbar, $b$) \\ kernel } & \multirowcell{2}{split dir.\ } & \multirowcell{2}{split dir.\ \\ kernel} \\
&  &  &  &  &  \\
\hline
\end{tabular}
\end{small}
\caption{Overview of the importance of tree parameters on performance indicators for the 8D data set. The more important a parameter, the higher up in the table it is, also within each row.}
\label{tab: importance parameters target quantities 8D}
\end{center}
\end{table}

\subsubsection{RMSE}

In these test runs we observe that the  $(\bar{N}, b)$ configuration has the strongest influence on the RMSE. Trees with the $(200,15)$ configuration have on average higher RMSE than those with the $(1000,1000)$ configuration.  The choice of split criterion is less influential, yet still important to the RMSE results. We find that selecting the splitting direction through the maximum spread per length scale yields the lowest RMSE on average, while splitting along the first principal component leads to the on average highest RMSE values. 

For the $(200,15)$ configuration we observe that the choice of kernel function is important, with the Matérn \nuMatern{3} kernel performing best, followed by the Matérn \nuMatern{5} kernel, and the Gaussian kernel yielding on average the highest RMSE. A maximum of 200 data points per GP is quite low when attempting to learn an 8D target function, even if each GP only deals with a local region of input space. It is therefore not surprising that the choice of kernel, which encodes the prior on function space, proves to be important.
We do not observe any notable impact on the RMSE from the overlap parameter \thetat.

\subsubsection{Fraction of test points with prediction error below 5\% (\pusable)}
In contrast to test setup 1, we observe no strong connection between tree hyperparameters' impact on the RMSE and their impact on \pusable. We attribute this to the small variance in \pusable for test setup 2. It is important to note that the largest and smallest values lie within 1.5\% of one another and $\delta_{0.05} > 0.985$. Therefore, this section concentrates on the more subtle differences.

Our findings show that the kernel type is the most influential parameter with regards to \pusable. The Gaussian kernel leads to the highest \pusable on average, while the Matérn kernel with \nuMatern{3} leads to the worst on average. 

Less important but still relevant are choices of $(\bar{N}, b)$ and \thetat. On the one hand, the combination $(200, 15)$ gives rise to trees with lower \pusable. On the other hand, \thetat = 0.1 leads to higher \pusable.
We find no noteworthy impact of the splitting criterion on \pusable.

\subsubsection{Prediction uncertainty}\label{sec: uncert est 8D}
The prediction uncertainty is most influenced by the splitting criterion, where \verb|max_spread_per_| \verb|lengthscale| leads to lower and \verb|principal_component| to higher uncertainty estimates. Slightly less important is the choice of overlap parameter \thetat. Here  the lowest estimates are obtained when gradual splitting is chosen. Conversely, larger \thetat yield higher average uncertainties. This is to be expected as we combine the predictions from an increasing number of local GPs as the overlap increases. 
The combination of \Nbar and $b$ as well as the choice of kernel have limited influence on the uncertainty estimate.

\subsubsection{Update time}

As discussed in Sec.\ \ref{sec: 4d training time}, the per-point update time for a tree is of course most strongly influenced by the choice of $(\bar{N}, b)$. Of the two configurations $(200,15)$ and $(1000,1000)$ that we consider for test setup 2, the latter leads to shorter update times. Also, as expected, using gradual splitting increases the update time.

Moreover, for the $(200,15)$ configuration, we see that using a Gaussian kernel more strongly contributes to a reduced training time. We find no relevant effect of the split direction criterion.

\subsubsection{Prediction time}

\Nbar and \thetat are the most important tree parameters affecting the per-point prediction time, as discussed in Sec.\ \ref{sec: 4d prediction time}, with higher \Nbar and \thetat values leading to longer prediction time. We notice this also in the runs with test setup 2. The slowest predictions are observed for the trees that use $(\bar{N}, b) = (1000,1000)$ and $\theta = 0.1$. 
The choice of split direction criterion and kernel type does not impact the prediction time, as expected.

\subsection{Learning from a uniformly-generated input stream}\label{sec: uniform input stream}

\begin{figure}\centering
\includegraphics[width=\textwidth]{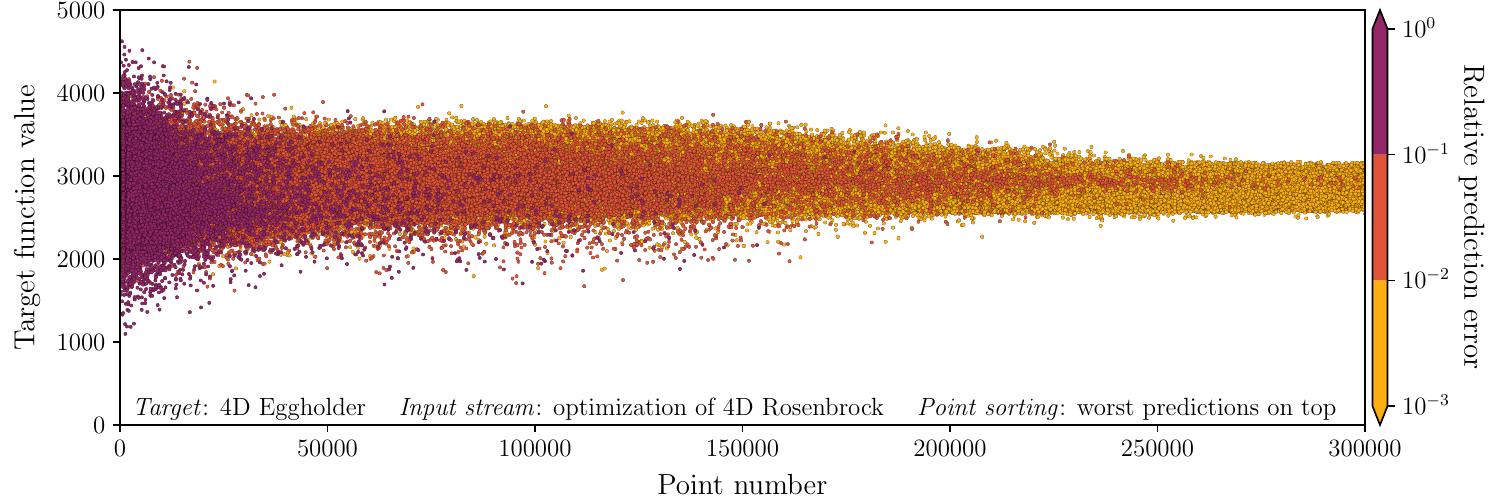}\vspace{0.3\baselineskip}
\includegraphics[width=\textwidth]{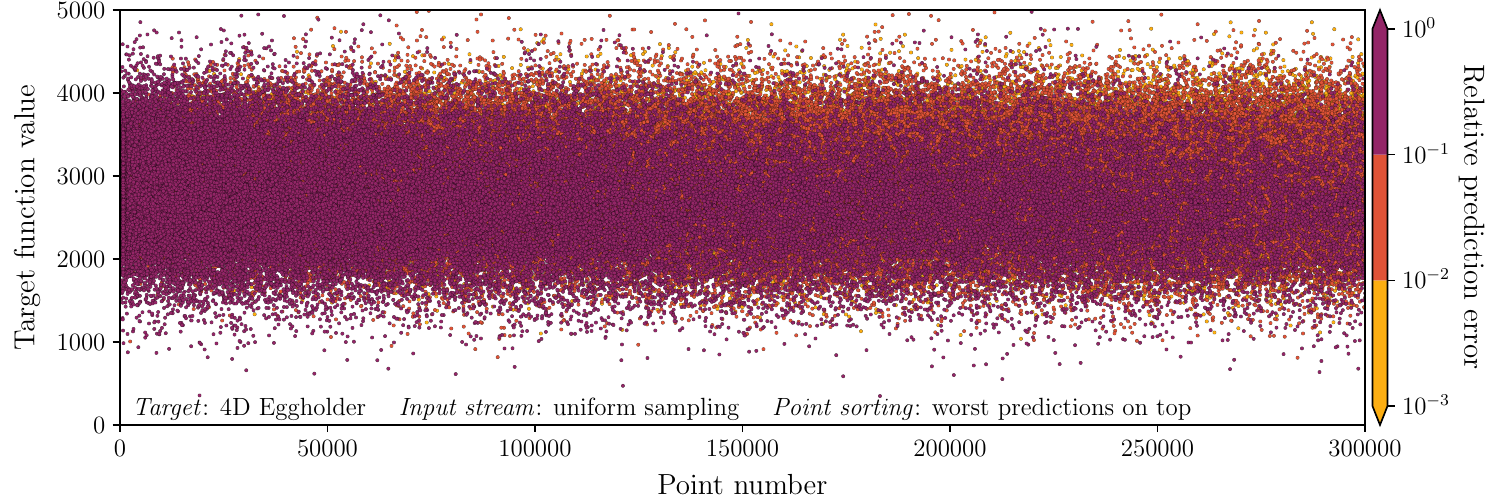}\vspace{0.3\baselineskip}
\includegraphics[width=\textwidth]{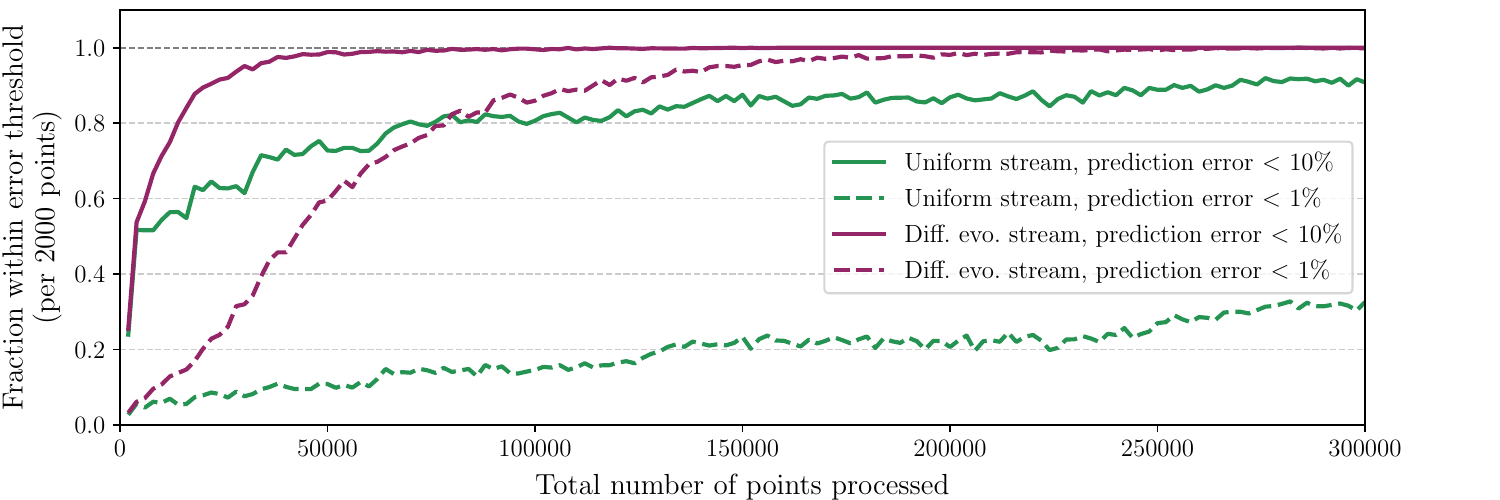}
\caption{\label{fig: test of uniform vs diff evo sampling} Illustration of how the nature of the input stream affects the challenge of learning a given target function. \textit{Top}: A \texttt{GPTreeO} test run with the four-dimensional Eggholder function as target function, where the stream of input points is generated through a differential evolution optimiser exploring a four-dimensional Rosenbrock function. \textit{Middle}: A similar test setup as for the top panel, but now the stream of input points is generated by uniform sampling from the full domain. The points in the scatter plots are sorted such that the points with largest prediction errors are displayed on top. \textit{Bottom}: The evolution of the prediction accuracy in the two test runs. The graphs show what fraction of predictions are within a given error threshold, computed for batches of 2000 input points. The green and purple graphs correspond to the test runs in the top and middle panels, respectively.}
\end{figure}

As discussed in Sec.\ \ref{sec: test setup}, the time evolution of \texttt{GPTreeO}'s prediction accuracy of course depends on the nature of the input data stream which \texttt{GPTreeO} passively receives. For instance, if the input points over time become more similar (as during a numerical optimisation), or if at least near-future input points are similar to already received points (as in a Markov Chain Monte Carlo), this will help \texttt{GPTreeO} to more quickly achieve low prediction errors. A particularly challenging case is therefore a scenario where the input points $\textbf{x}$ are always drawn uniformly from the entire input domain. 

We illustrate this in Figure \ref{fig: test of uniform vs diff evo sampling} with results from two \texttt{GPTreeO} test runs, where the target function is the four-dimensional Eggholder function. The top panel shows all the true function values and \texttt{GPTreeO}'s relative prediction errors when the input points are sampled by a differential evolution optimiser exploring the four-dimensional Rosenbrock function. The points in the plots are sorted such that points with large prediction error are displayed on top.
In contrast, the middle panel shows the corresponding result from a run where the input points are uniformly sampled from the full domain. The bottom panel compares the evolution of \texttt{GPTreeO}'s prediction accuracy in the two test runs. For batches of 2000 input points, the graphs show what fraction of predictions are within 10\% (solid lines) or 1\% (dashed lines) of the true function value. In both test runs the prediction accuracy gradually increases as a function of the number of processed points, but, as expected, the prediction accuracy improves much more slowly for the run with the uniformly-generated input stream. 

\subsection{Learning a noisy target function}\label{sec: noisy target function}
In many realistic use-cases the input stream will consist of noisy data. We here briefly investigate how such noise can affect the relative importance of the different tree hyperparameters for \texttt{GPTreeO}'s performance. For this purpose we repeat all the test runs with $(\bar{N}, b) = (1000,1000)$ from test setup 1, but now add 1\% Gaussian noise to the target variable. Below we focus on the impact on the RMSE. 
The complete set of results from noisy runs are given in Table \ref{tab: results trees 13000000} in the Appendix, and the noise-free ones can be found in Table \ref{tab: results trees 12000000}.

As expected, the lowest RMSE values in the comparison are achieved by trees trained on noise-free data. Yet, the worst RMSE in the noisy and noise-free case are comparable. Similar to Sec.\ \ref{sec: RMSE 4D}, the choice of kernel has the biggest impact on RMSE, where the Gaussian kernel increases the RMSE and a Matérn kernel with \nuMatern{3} reduces it. There is no noticeable effect in changing the overlap parameter or the splitting criterion.

\section{Conclusion}\label{sec: conclusions}

We have introduced the \texttt{R} package \texttt{GPTreeO}, which implements a divide-and-conquer approach to Gaussian processes through local approximations in the context of online (machine) learning. 
Built on the Dividing Local Gaussian Processes approach, the algorithm extensions and modularity introduced in \texttt{GPTreeO} ensures that this approach can be adapted to applications with a wide range of requirements on computational speed, regression accuracy, uncertainty estimation, and the stability and smoothness of predictions. In particular, \texttt{GPTreeO} enables fine-grained control of how frequently local GPs are retrained, on-the-fly uncertainty calibration, and introduces new methods to control how the training data is partitioned as the GP tree grows.

To provide guidance on how to best configure \texttt{GPTreeO}, we studied the impact of each algorithm feature in detail for two test cases, focusing on the effect on prediction accuracy, uncertainty estimation and computational speed. For example, we showed how \texttt{GPTreeO}'s prediction accuracy is most strongly influenced by the choice of size cap, retrain frequency and kernel type for the local GPs, with some influence also from the degree of overlap between neighbouring GPs.

\texttt{GPTreeO} organises the local GPs in a single tree and uses the resulting model for prediction. It may be advantageous to extend the approach towards an ensemble of trees, mimicking the strategies of bagging and random forests. Here we imagine the prediction based on a set of different trees, trained in slightly different ways on the same data stream (see e.g.\ \cite{CaoAl2015}, \cite{LiuAl2018}) or trained similarly on small modifications of the data. This is material for future work. Other possible extensions to the algorithm include an adaptive and automatic tuning of tree hyperparameters, to reduce the number of parameters the user must set, and a ``continual forgetting'' option, that automatically removes the least relevant points (e.g.\ the oldest points, or the points least informative for recent predictions) to keep the model size constant. 

This first version of \texttt{GPTreeO} is implemented as an \texttt{R} package to make it practical for the statistical community, but we plan to also release a \texttt{Python} version in the near future.

\section*{Acknowledgments}

We thank our colleagues in the PLUMBIN' research project for helpful discussions. AK wishes to thank Erik Alexander Sandvik for many useful discussions on Gaussian processes and local approximations, and Pat Scott, Ben Farmer, Nicholas Reed, and Iza Veli{\v s}{\v c}ek for earlier collaboration on this topic. This work has been supported by the Research Council of Norway (RCN) through the FRIPRO grant 323985 PLUMBIN’. 

\section*{Appendix}\label{appendix}

In this appendix we provide the results for all performance indicators on all our test runs. Each row in each table contains the tree parameters for a specific tree and the performance indicator results achieved in the given run. The performance indicators are introduced in Sec.\ \ref{sec: sensitivity analysis} and the exact test setups are described in detail in Sec. \ref{sec: test setup}. Tables \ref{tab: results trees 8000000}--\ref{tab: results trees 12000000} show the results with test setup 1, where we train on a four-dimensional input stream. The corresponding performance indicators are analysed in Sec.\ \ref{sec: 4D test results}. Similarly, Tables \ref{tab: results trees 15000000} and \ref{tab: results trees 7000000} contain the results from runs with test setup 2. Here the data stream has eight input dimensions, and the performance indicator results are discussed in Sec. \ref{sec: 8D test results}. Lastly, Table \ref{tab: results trees 13000000} contains the results from test runs similar to those in Table \ref{tab: results trees 12000000}, except that 1\% noise is added to the target variable. These results are discussed in Sec.\ \ref{sec: noisy target function}.

In the following tables we introduce several short-hand notations: we abbreviate the term \textit{splitting criteria} as ``Split.\ crit.'', while ``max sprd'' and ``max sprd/$l$'' refer to the options \verb|"max_spread"| and \verb|"max_spread_per_lengthscale"|, respectively. Similarly, ``principal'' stands for \verb|principal_| \verb|component|. The updating time is denoted $t_\text{upd}$, and we have also dropped the leading 0's in the \pusable, $t_\text{upd}$ and $t_\text{pred}$ columns.

\newpage
\begin{longtable}{|c|c|c|c|c|c|c|c|c|}
\hline
\endfoot
\caption{Table used in Sec.\ \ref{sec: 4D test results} (test setup 1). This table contains the runs from \Nbar = 100, retrain buffer length = 2} \\
    \hline
    \Nbar; $b$ & $\theta$ & Kernel & Split.\ crit.\ & RMSE & \pusable & Uncert.\ & $t_\text{upd}$ & $t_\text{pred}$
    \begin{filecontents*}{all_trees_8000000.csv}
lumber,Identifier,ThetaPara,KernelType,SplitDir,RMSETarg,pUsable,yErrUnscal,timeTrainAvg,timePredAvg
1,(100; 2),0.01,Gauss,max sprd,227.313,.913,162.356,.030,.0015
2,(100; 2),0.05,Gauss,max sprd,225.119,.931,180.689,.030,.0021
3,(100; 2),0.10,Gauss,max sprd,222.176,.926,202.083,.029,.0033
4,(100; 2),grad.\ split,Gauss,max sprd,221.166,.925,156.792,.043,.0014
5,(100; 2),0.01,Mat 3/2,max sprd,219.144,.923,162.994,.030,.0015
6,(100; 2),0.05,Mat 3/2,max sprd,215.286,.934,177.301,.030,.0021
7,(100; 2),0.10,Mat 3/2,max sprd,215.596,.937,198.175,.032,.0035
8,(100; 2),grad.\ split,Mat 3/2,max sprd,211.195,.941,155.597,.042,.0014
9,(100; 2),0.01,Mat 5/2,max sprd,221.703,.922,161.465,.030,.0014
10,(100; 2),0.05,Mat 5/2,max sprd,216.181,.935,175.842,.032,.0021
11,(100; 2),0.10,Mat 5/2,max sprd,215.424,.932,197.250,.033,.0035
12,(100; 2),grad.\ split,Mat 5/2,max sprd,213.568,.932,154.511,.043,.0013
13,(100; 2),0.01,Gauss,max sprd/$l$,229.217,.932,164.663,.034,.0016
14,(100; 2),0.05,Gauss,max sprd/$l$,222.690,.922,182.128,.029,.0022
15,(100; 2),0.10,Gauss,max sprd/$l$,225.200,.923,208.733,.029,.0037
16,(100; 2),grad.\ split,Gauss,max sprd/$l$,222.624,.921,156.951,.039,.0013
17,(100; 2),0.01,Mat 3/2,max sprd/$l$,218.992,.931,163.834,.035,.0017
18,(100; 2),0.05,Mat 3/2,max sprd/$l$,216.501,.937,181.431,.032,.0024
19,(100; 2),0.10,Mat 3/2,max sprd/$l$,215.033,.939,201.186,.033,.0040
20,(100; 2),grad.\ split,Mat 3/2,max sprd/$l$,207.890,.940,152.434,.043,.0014
21,(100; 2),0.01,Mat 5/2,max sprd/$l$,221.758,.932,162.073,.032,.0015
22,(100; 2),0.05,Mat 5/2,max sprd/$l$,215.482,.950,178.355,.032,.0023
23,(100; 2),0.10,Mat 5/2,max sprd/$l$,215.317,.943,201.117,.033,.0038
24,(100; 2),grad.\ split,Mat 5/2,max sprd/$l$,213.710,.949,153.393,.043,.0013
25,(100; 2),0.01,Gauss,principal,226.899,.918,167.368,.029,.0017
26,(100; 2),0.05,Gauss,principal,222.455,.924,191.559,.029,.0028
27,(100; 2),0.10,Gauss,principal,223.003,.915,221.953,.030,.0051
28,(100; 2),grad.\ split,Gauss,principal,223.348,.919,159.883,.044,.0016
29,(100; 2),0.01,Mat 3/2,principal,220.125,.918,170.150,.033,.0019
30,(100; 2),0.05,Mat 3/2,principal,216.296,.932,191.409,.034,.0032
31,(100; 2),0.10,Mat 3/2,principal,217.298,.908,219.708,.035,.0059
32,(100; 2),grad.\ split,Mat 3/2,principal,215.624,.914,162.974,.042,.0015
33,(100; 2),0.01,Mat 5/2,principal,222.076,.918,167.688,.034,.0018
34,(100; 2),0.05,Mat 5/2,principal,217.958,.927,190.564,.034,.0031
35,(100; 2),0.10,Mat 5/2,principal,217.801,.917,218.557,.031,.0049
36,(100; 2),grad.\ split,Mat 5/2,principal,216.871,.920,160.136,.037,.0012
\end{filecontents*}
\csvreader[head to column names]{all_trees_8000000.csv}{}
    {\\\hline \Identifier & \ThetaPara & \KernelType & \SplitDir & \RMSETarg & \pUsable & \yErrUnscal & \timeTrainAvg & \timePredAvg}
    \label{tab: results trees 8000000}
\end{longtable}

\newpage
\begin{longtable}{|c|c|c|c|c|c|c|c|c|}
\hline
\endfoot
\caption{Table used in Sec.\ \ref{sec: 4D test results} (test setup 1). This table contains the runs from \Nbar = 200, retrain buffer length = 15} \\
    \hline
    \Nbar; $b$ & $\theta$ & Kernel & Split.\ crit.\ & RMSE & \pusable & Uncert.\ & $t_\text{upd}$ & $t_\text{pred}$
    \begin{filecontents*}{all_trees_11000000.csv}
lumber,Identifier,ThetaPara,KernelType,SplitDir,RMSETarg,pUsable,yErrUnscal,timeTrainAvg,timePredAvg
37,(200; 15),0.01,Gauss,max sprd,217.702,.931,166.786,.013,.0014
38,(200; 15),0.05,Gauss,max sprd,214.342,.936,180.310,.012,.0020
39,(200; 15),0.10,Gauss,max sprd,213.149,.932,203.532,.014,.0034
40,(200; 15),grad.\ split,Gauss,max sprd,215.668,.931,162.696,.020,.0014
41,(200; 15),0.01,Mat 3/2,max sprd,207.237,.944,162.383,.013,.0014
42,(200; 15),0.05,Mat 3/2,max sprd,204.723,.948,175.165,.014,.0022
43,(200; 15),0.10,Mat 3/2,max sprd,205.710,.946,196.388,.015,.0038
44,(200; 15),grad.\ split,Mat 3/2,max sprd,203.529,.948,156.230,.020,.0014
45,(200; 15),0.01,Mat 5/2,max sprd,210.372,.943,162.784,.013,.0014
46,(200; 15),0.05,Mat 5/2,max sprd,207.160,.944,175.616,.015,.0023
47,(200; 15),0.10,Mat 5/2,max sprd,207.143,.939,197.648,.015,.0038
48,(200; 15),grad.\ split,Mat 5/2,max sprd,207.093,.947,157.199,.022,.0014
49,(200; 15),0.01,Gauss,max sprd/$l$,216.222,.941,165.516,.012,.0014
50,(200; 15),0.05,Gauss,max sprd/$l$,212.987,.932,178.907,.012,.0020
51,(200; 15),0.10,Gauss,max sprd/$l$,211.823,.938,201.507,.015,.0038
52,(200; 15),grad.\ split,Gauss,max sprd/$l$,213.361,.943,161.131,.018,.0013
53,(200; 15),0.01,Mat 3/2,max sprd/$l$,206.415,.949,159.675,.014,.0016
54,(200; 15),0.05,Mat 3/2,max sprd/$l$,204.910,.956,174.230,.015,.0025
55,(200; 15),0.10,Mat 3/2,max sprd/$l$,204.554,.949,195.093,.014,.0039
56,(200; 15),grad.\ split,Mat 3/2,max sprd/$l$,202.854,.955,153.402,.020,.0015
57,(200; 15),0.01,Mat 5/2,max sprd/$l$,208.035,.942,159.499,.014,.0015
58,(200; 15),0.05,Mat 5/2,max sprd/$l$,204.703,.950,173.184,.016,.0024
59,(200; 15),0.10,Mat 5/2,max sprd/$l$,206.547,.955,193.666,.014,.0033
60,(200; 15),grad.\ split,Mat 5/2,max sprd/$l$,205.261,.949,155.982,.020,.0014
61,(200; 15),0.01,Gauss,principal,217.161,.936,171.429,.014,.0018
62,(200; 15),0.05,Gauss,principal,214.098,.939,198.595,.014,.0031
63,(200; 15),0.10,Gauss,principal,214.151,.920,230.900,.016,.0065
64,(200; 15),grad.\ split,Gauss,principal,216.551,.927,166.093,.019,.0015
65,(200; 15),0.01,Mat 3/2,principal,210.002,.946,170.511,.015,.0019
66,(200; 15),0.05,Mat 3/2,principal,206.946,.944,193.231,.014,.0030
67,(200; 15),0.10,Mat 3/2,principal,207.622,.922,221.887,.014,.0056
68,(200; 15),grad.\ split,Mat 3/2,principal,206.861,.944,163.456,.018,.0014
69,(200; 15),0.01,Mat 5/2,principal,212.110,.944,169.338,.015,.0017
70,(200; 15),0.05,Mat 5/2,principal,207.850,.951,193.644,.014,.0028
71,(200; 15),0.10,Mat 5/2,principal,209.010,.921,223.977,.012,.0047
72,(200; 15),grad.\ split,Mat 5/2,principal,210.351,.946,163.126,.015,.0010
\end{filecontents*}
\csvreader[head to column names]{all_trees_11000000.csv}{} 
    {\\\hline \Identifier & \ThetaPara & \KernelType & \SplitDir & \RMSETarg & \pUsable & \yErrUnscal & \timeTrainAvg & \timePredAvg}
    \label{tab: results trees 11000000}
\end{longtable}

\newpage
\begin{longtable}{|c|c|c|c|c|c|c|c|c|}
\hline
\endfoot
\caption{Table used in Sec.\ \ref{sec: 4D test results} (test setup 1). This table contains the runs from \Nbar = 1000, retrain buffer length = 1000. It also used in Sec.\ \ref{sec: noisy target function}. It does not contain any noise} \\
    \hline
    \Nbar; $b$ & $\theta$ & Kernel & Split.\ crit.\ & RMSE & \pusable & Uncert.\ & $t_\text{upd}$ & $t_\text{pred}$
    \begin{filecontents*}{all_trees_12000000.csv}
lumber,Identifier,ThetaPara,KernelType,SplitDir,RMSETarg,pUsable,yErrUnscal,timeTrainAvg,timePredAvg
73,(1000; 1000),0.01,Gauss,max sprd,228.207,.906,200.412,.010,.0031
74,(1000; 1000),0.05,Gauss,max sprd,225.044,.931,217.834,.011,.0046
75,(1000; 1000),0.10,Gauss,max sprd,224.292,.915,242.379,.011,.0074
76,(1000; 1000),grad.\ split,Gauss,max sprd,228.049,.900,197.673,.029,.0059
77,(1000; 1000),0.01,Mat 3/2,max sprd,213.270,.925,182.465,.008,.0030
78,(1000; 1000),0.05,Mat 3/2,max sprd,211.401,.941,196.806,.008,.0045
79,(1000; 1000),0.10,Mat 3/2,max sprd,210.614,.934,215.970,.009,.0075
80,(1000; 1000),grad.\ split,Mat 3/2,max sprd,209.512,.934,173.771,.018,.0058
81,(1000; 1000),0.01,Mat 5/2,max sprd,217.810,.923,187.352,.009,.0030
82,(1000; 1000),0.05,Mat 5/2,max sprd,215.720,.940,202.705,.010,.0049
83,(1000; 1000),0.10,Mat 5/2,max sprd,214.675,.930,223.774,.009,.0079
84,(1000; 1000),grad.\ split,Mat 5/2,max sprd,215.279,.922,180.094,.020,.0061
85,(1000; 1000),0.01,Gauss,max sprd/$l$,225.326,.910,193.277,.011,.0033
86,(1000; 1000),0.05,Gauss,max sprd/$l$,222.888,.922,211.619,.013,.0051
87,(1000; 1000),0.10,Gauss,max sprd/$l$,222.064,.911,234.339,.011,.0078
88,(1000; 1000),grad.\ split,Gauss,max sprd/$l$,227.329,.911,193.516,.035,.0062
89,(1000; 1000),0.01,Mat 3/2,max sprd/$l$,209.277,.932,173.264,.010,.0034
90,(1000; 1000),0.05,Mat 3/2,max sprd/$l$,207.167,.933,187.032,.010,.0052
91,(1000; 1000),0.10,Mat 3/2,max sprd/$l$,207.978,.929,208.442,.009,.0085
92,(1000; 1000),grad.\ split,Mat 3/2,max sprd/$l$,208.617,.940,166.824,.018,.0062
93,(1000; 1000),0.01,Mat 5/2,max sprd/$l$,215.244,.920,179.739,.011,.0036
94,(1000; 1000),0.05,Mat 5/2,max sprd/$l$,212.304,.936,194.523,.010,.0054
95,(1000; 1000),0.10,Mat 5/2,max sprd/$l$,212.672,.926,217.404,.010,.0088
96,(1000; 1000),grad.\ split,Mat 5/2,max sprd/$l$,214.873,.938,174.857,.020,.0064
97,(1000; 1000),0.01,Gauss,principal,228.023,.914,204.342,.012,.0038
98,(1000; 1000),0.05,Gauss,principal,222.440,.921,231.210,.011,.0067
99,(1000; 1000),0.10,Gauss,principal,223.033,.914,270.361,.011,.0098
100,(1000; 1000),grad.\ split,Gauss,principal,229.903,.903,200.368,.019,.0044
101,(1000; 1000),0.01,Mat 3/2,principal,213.081,.927,189.822,.009,.0037
102,(1000; 1000),0.05,Mat 3/2,principal,209.036,.940,210.693,.009,.0059
103,(1000; 1000),0.10,Mat 3/2,principal,212.046,.919,243.954,.008,.0104
104,(1000; 1000),grad.\ split,Mat 3/2,principal,212.045,.920,180.430,.015,.0049
105,(1000; 1000),0.01,Mat 5/2,principal,217.825,.917,193.075,.009,.0033
106,(1000; 1000),0.05,Mat 5/2,principal,213.027,.936,215.985,.009,.0058
107,(1000; 1000),0.10,Mat 5/2,principal,215.196,.925,252.028,.008,.0105
108,(1000; 1000),grad.\ split,Mat 5/2,principal,217.560,.919,185.009,.016,.0050
\end{filecontents*}
\csvreader[head to column names]{all_trees_12000000.csv}{} 
    {\\\hline \Identifier & \ThetaPara & \KernelType & \SplitDir & \RMSETarg & \pUsable & \yErrUnscal & \timeTrainAvg & \timePredAvg}
    \label{tab: results trees 12000000}
\end{longtable}

\newpage
\begin{longtable}{|c|c|c|c|c|c|c|c|c|}
\hline
\endfoot
\caption{Table used in Sec.\ \ref{sec: 8D test results} (test setup 2). This table contains the runs from \Nbar = 200, retrain buffer length = 15} \\
    \hline
    \Nbar; $b$ & $\theta$ & Kernel & Split.\ crit.\ & RMSE & \pusable & Uncert.\ & $t_\text{upd}$ & $t_\text{pred}$
    \begin{filecontents*}{all_trees_15000000.csv}
lumber,Identifier,ThetaPara,KernelType,SplitDir,RMSETarg,pUsable,yErrUnscal,timeTrainAvg,timePredAvg
109,(200; 15),0.01,Gauss,max sprd,0.098,.994,0.051,.041,.0016
110,(200; 15),0.05,Gauss,max sprd,0.093,.995,0.057,.046,.0027
111,(200; 15),0.10,Gauss,max sprd,0.089,.996,0.064,.045,.0050
112,(200; 15),grad.\ split,Gauss,max sprd,0.088,.993,0.042,.064,.0015
113,(200; 15),0.01,Mat 3/2,max sprd,0.083,.986,0.058,.059,.0017
114,(200; 15),0.05,Mat 3/2,max sprd,0.077,.991,0.065,.059,.0028
115,(200; 15),0.10,Mat 3/2,max sprd,0.074,.992,0.074,.056,.0048
116,(200; 15),grad.\ split,Mat 3/2,max sprd,0.073,.991,0.050,.083,.0015
117,(200; 15),0.01,Mat 5/2,max sprd,0.092,.992,0.053,.053,.0016
118,(200; 15),0.05,Mat 5/2,max sprd,0.087,.993,0.060,.057,.0027
119,(200; 15),0.10,Mat 5/2,max sprd,0.088,.993,0.069,.056,.0049
120,(200; 15),grad.\ split,Mat 5/2,max sprd,0.084,.995,0.046,.083,.0015
121,(200; 15),0.01,Gauss,max sprd/$l$,0.094,.993,0.044,.040,.0017
122,(200; 15),0.05,Gauss,max sprd/$l$,0.092,.994,0.051,.038,.0025
123,(200; 15),0.10,Gauss,max sprd/$l$,0.088,.999,0.056,.041,.0046
124,(200; 15),grad.\ split,Gauss,max sprd/$l$,0.086,.992,0.038,.057,.0015
125,(200; 15),0.01,Mat 3/2,max sprd/$l$,0.059,.994,0.039,.050,.0016
126,(200; 15),0.05,Mat 3/2,max sprd/$l$,0.056,.991,0.044,.048,.0024
127,(200; 15),0.10,Mat 3/2,max sprd/$l$,0.054,.997,0.050,.052,.0042
128,(200; 15),grad.\ split,Mat 3/2,max sprd/$l$,0.054,.990,0.034,.075,.0015
129,(200; 15),0.01,Mat 5/2,max sprd/$l$,0.087,.994,0.047,.048,.0016
130,(200; 15),0.05,Mat 5/2,max sprd/$l$,0.085,.997,0.052,.049,.0025
131,(200; 15),0.10,Mat 5/2,max sprd/$l$,0.084,.993,0.061,.052,.0048
132,(200; 15),grad.\ split,Mat 5/2,max sprd/$l$,0.081,.997,0.041,.076,.0015
133,(200; 15),0.01,Gauss,principal,0.108,.995,0.056,.044,.0019
134,(200; 15),0.05,Gauss,principal,0.106,.991,0.068,.044,.0038
135,(200; 15),0.10,Gauss,principal,0.099,.997,0.083,.046,.0088
136,(200; 15),grad.\ split,Gauss,principal,0.102,.991,0.048,.065,.0016
137,(200; 15),0.01,Mat 3/2,principal,0.095,.991,0.065,.058,.0020
138,(200; 15),0.05,Mat 3/2,principal,0.086,.989,0.075,.059,.0039
139,(200; 15),0.10,Mat 3/2,principal,0.081,.993,0.089,.056,.0080
140,(200; 15),grad.\ split,Mat 3/2,principal,0.087,.989,0.057,.074,.0014
141,(200; 15),0.01,Mat 5/2,principal,0.101,.993,0.059,.051,.0017
142,(200; 15),0.05,Mat 5/2,principal,0.099,.991,0.071,.050,.0033
143,(200; 15),0.10,Mat 5/2,principal,0.092,.996,0.084,.047,.0063
144,(200; 15),grad.\ split,Mat 5/2,principal,0.098,.989,0.052,.058,.0011
\end{filecontents*}
\csvreader[head to column names]{all_trees_15000000.csv}{} 
    {\\\hline \Identifier & \ThetaPara & \KernelType & \SplitDir & \RMSETarg & \pUsable & \yErrUnscal & \timeTrainAvg & \timePredAvg}
    \label{tab: results trees 15000000}
\end{longtable}

\newpage
\begin{longtable}{|c|c|c|c|c|c|c|c|c|}
\hline
\endfoot
\caption{Table used in Sec.\ \ref{sec: 8D test results} (test setup 2). This table contains the runs from \Nbar = 1000, retrain buffer length = 1000} \\
    \hline
    \Nbar; $b$ & $\theta$ & Kernel & Split.\ crit.\ & RMSE & \pusable & Uncert.\ & $t_\text{upd}$ & $t_\text{pred}$
    \begin{filecontents*}{all_trees_7000000.csv}
lumber,Identifier,ThetaPara,KernelType,SplitDir,RMSETarg,pUsable,yErrUnscal,timeTrainAvg,timePredAvg
145,(1000; 1000),0.01,Gauss,max sprd,0.057,.996,0.034,.026,.0036
146,(1000; 1000),0.05,Gauss,max sprd,0.054,.997,0.038,.023,.0054
147,(1000; 1000),0.10,Gauss,max sprd,0.053,.997,0.044,.027,.0096
148,(1000; 1000),grad.\ split,Gauss,max sprd,0.050,.998,0.028,.051,.0061
149,(1000; 1000),0.01,Mat 3/2,max sprd,0.065,.990,0.048,.029,.0036
150,(1000; 1000),0.05,Mat 3/2,max sprd,0.061,.991,0.052,.026,.0055
151,(1000; 1000),0.10,Mat 3/2,max sprd,0.060,.992,0.058,.031,.0097
152,(1000; 1000),grad.\ split,Mat 3/2,max sprd,0.056,.992,0.039,.064,.0064
153,(1000; 1000),0.01,Mat 5/2,max sprd,0.059,.992,0.039,.028,.0037
154,(1000; 1000),0.05,Mat 5/2,max sprd,0.055,.995,0.043,.028,.0057
155,(1000; 1000),0.10,Mat 5/2,max sprd,0.055,.996,0.049,.033,.0099
156,(1000; 1000),grad.\ split,Mat 5/2,max sprd,0.051,.996,0.032,.058,.0063
157,(1000; 1000),0.01,Gauss,max sprd/$l$,0.048,.993,0.026,.023,.0036
158,(1000; 1000),0.05,Gauss,max sprd/$l$,0.046,.995,0.029,.022,.0051
159,(1000; 1000),0.10,Gauss,max sprd/$l$,0.044,.998,0.033,.025,.0077
160,(1000; 1000),grad.\ split,Gauss,max sprd/$l$,0.046,.998,0.024,.042,.0061
161,(1000; 1000),0.01,Mat 3/2,max sprd/$l$,0.054,.995,0.037,.028,.0035
162,(1000; 1000),0.05,Mat 3/2,max sprd/$l$,0.052,.994,0.040,.029,.0052
163,(1000; 1000),0.10,Mat 3/2,max sprd/$l$,0.052,.992,0.044,.025,.0077
164,(1000; 1000),grad.\ split,Mat 3/2,max sprd/$l$,0.050,.996,0.033,.054,.0060
165,(1000; 1000),0.01,Mat 5/2,max sprd/$l$,0.049,.996,0.030,.027,.0035
166,(1000; 1000),0.05,Mat 5/2,max sprd/$l$,0.047,.997,0.033,.025,.0052
167,(1000; 1000),0.10,Mat 5/2,max sprd/$l$,0.047,.996,0.038,.025,.0083
168,(1000; 1000),grad.\ split,Mat 5/2,max sprd/$l$,0.046,.998,0.027,.053,.0061
169,(1000; 1000),0.01,Gauss,principal,0.067,.995,0.041,.029,.0037
170,(1000; 1000),0.05,Gauss,principal,0.063,.996,0.049,.025,.0074
171,(1000; 1000),0.10,Gauss,principal,0.059,.995,0.056,.028,.0149
172,(1000; 1000),grad.\ split,Gauss,principal,0.057,.993,0.033,.048,.0058
173,(1000; 1000),0.01,Mat 3/2,principal,0.077,.995,0.057,.032,.0039
174,(1000; 1000),0.05,Mat 3/2,principal,0.073,.992,0.064,.029,.0078
175,(1000; 1000),0.10,Mat 3/2,principal,0.069,.994,0.072,.030,.0147
176,(1000; 1000),grad.\ split,Mat 3/2,principal,0.064,.995,0.046,.049,.0054
177,(1000; 1000),0.01,Mat 5/2,principal,0.070,.995,0.047,.032,.0039
178,(1000; 1000),0.05,Mat 5/2,principal,0.067,.995,0.054,.025,.0068
179,(1000; 1000),0.10,Mat 5/2,principal,0.062,.995,0.062,.023,.1070
180,(1000; 1000),grad.\ split,Mat 5/2,principal,0.058,.993,0.037,.035,.0041
\end{filecontents*}
\csvreader[head to column names]{all_trees_7000000.csv}{} 
    {\\\hline \Identifier & \ThetaPara & \KernelType & \SplitDir & \RMSETarg & \pUsable & \yErrUnscal & \timeTrainAvg & \timePredAvg}
    \label{tab: results trees 7000000}
\end{longtable}

\newpage
\begin{longtable}{|c|c|c|c|c|c|c|c|c|}
\hline
\endfoot
\caption{Table used in Sec.\ \ref{sec: noisy target function}. In contrast to trees shown in Tab. \ref{tab: results trees 12000000}, 1\% noise was added to the training data.} \\
    \hline
    \Nbar; $b$ & $\theta$ & Kernel & Split.\ crit.\ & RMSE & \pusable & Uncert.\ & $t_\text{upd}$ & $t_\text{pred}$
    \begin{filecontents*}{all_trees_13000000.csv}
lumber,Identifier,ThetaPara,KernelType,SplitDir,RMSETarg,pUsable,yErrUnscal,timeTrainAvg,timePredAvg
181,(1000; 1000),0.01,Gauss,max sprd,230.470,.907,192.130,.009,.0031
182,(1000; 1000),0.05,Gauss,max sprd,229.414,.904,209.510,.011,.0047
183,(1000; 1000),0.10,Gauss,max sprd,226.150,.912,233.395,.011,.0076
184,(1000; 1000),grad.\ split,Gauss,max sprd,228.857,.900,185.577,.021,.0058
185,(1000; 1000),0.01,Mat 3/2,max sprd,220.051,.919,184.982,.009,.0031
186,(1000; 1000),0.05,Mat 3/2,max sprd,218.177,.910,199.181,.010,.0047
187,(1000; 1000),0.10,Mat 3/2,max sprd,218.106,.928,219.597,.010,.0076
188,(1000; 1000),grad.\ split,Mat 3/2,max sprd,216.524,.922,176.814,.020,.0061
189,(1000; 1000),0.01,Mat 5/2,max sprd,224.629,.921,186.671,.010,.0032
190,(1000; 1000),0.05,Mat 5/2,max sprd,220.653,.907,201.069,.010,.0050
191,(1000; 1000),0.10,Mat 5/2,max sprd,219.187,.925,221.747,.011,.0082
192,(1000; 1000),grad.\ split,Mat 5/2,max sprd,218.178,.915,176.994,.019,.0059
193,(1000; 1000),0.01,Gauss,max sprd/$l$,230.606,.913,192.586,.009,.0032
194,(1000; 1000),0.05,Gauss,max sprd/$l$,226.593,.915,206.154,.009,.0046
195,(1000; 1000),0.10,Gauss,max sprd/$l$,224.650,.894,227.939,.011,.0074
196,(1000; 1000),grad.\ split,Gauss,max sprd/$l$,230.008,.883,187.723,.019,.0057
197,(1000; 1000),0.01,Mat 3/2,max sprd/$l$,217.398,.917,181.984,.009,.0033
198,(1000; 1000),0.05,Mat 3/2,max sprd/$l$,216.311,.917,196.396,.010,.0051
199,(1000; 1000),0.10,Mat 3/2,max sprd/$l$,216.074,.920,217.555,.010,.0080
200,(1000; 1000),grad.\ split,Mat 3/2,max sprd/$l$,216.341,.918,170.094,.020,.0059
201,(1000; 1000),0.01,Mat 5/2,max sprd/$l$,219.529,.923,179.437,.010,.0033
202,(1000; 1000),0.05,Mat 5/2,max sprd/$l$,217.262,.931,195.565,.010,.0047
203,(1000; 1000),0.10,Mat 5/2,max sprd/$l$,216.630,.910,216.759,.010,.0072
204,(1000; 1000),grad.\ split,Mat 5/2,max sprd/$l$,218.157,.923,176.350,.018,.0059
205,(1000; 1000),0.01,Gauss,principal,226.635,.885,194.674,.013,.0036
206,(1000; 1000),0.05,Gauss,principal,223.292,.907,218.941,.009,.0061
207,(1000; 1000),0.10,Gauss,principal,224.647,.906,256.565,.010,.0116
208,(1000; 1000),grad.\ split,Gauss,principal,229.672,.886,189.551,.021,.0057
209,(1000; 1000),0.01,Mat 3/2,principal,216.546,.914,191.517,.010,.0036
210,(1000; 1000),0.05,Mat 3/2,principal,214.360,.910,209.508,.010,.0063
211,(1000; 1000),0.10,Mat 3/2,principal,216.679,.906,241.670,.009,.0106
212,(1000; 1000),grad.\ split,Mat 3/2,principal,218.266,.896,182.318,.016,.0051
213,(1000; 1000),0.01,Mat 5/2,principal,218.036,.903,189.954,.009,.0034
214,(1000; 1000),0.05,Mat 5/2,principal,215.060,.915,211.032,.009,.0056
215,(1000; 1000),0.10,Mat 5/2,principal,217.113,.907,245.033,.008,.0093
216,(1000; 1000),grad.\ split,Mat 5/2,principal,219.853,.905,182.572,.013,.0041
\end{filecontents*}
\csvreader[head to column names]{all_trees_13000000.csv}{} 
    {\\\hline \Identifier & \ThetaPara & \KernelType & \SplitDir & \RMSETarg & \pUsable & \yErrUnscal & \timeTrainAvg & \timePredAvg}
    \label{tab: results trees 13000000}
\end{longtable}

\bibliographystyle{jss}
\bibliography{biblio}

\end{document}